\documentclass[10pt,twocolumn,letterpaper]{article}

\usepackage[pagenumbers]{cvpr} % To force page numbers, e.g. for an arXiv version

\usepackage{graphicx}
\usepackage{amsmath}
\usepackage{amssymb}
\usepackage{booktabs}

\usepackage[pagebackref,breaklinks,colorlinks]{hyperref}

\usepackage[capitalize]{cleveref}
\crefname{section}{Sec.}{Secs.}
\Crefname{section}{Section}{Sections}
\Crefname{table}{Table}{Tables}
\crefname{table}{Tab.}{Tabs.}

\usepackage{amsfonts}       % blackboard math symbols
\usepackage{nicefrac}       % compact symbols for 1/2, etc.
\usepackage{microtype}      % microtypography
\usepackage{xcolor}         % colors

\usepackage{times}
\usepackage{epsfig}
\usepackage{graphicx}
\usepackage{mathtools}
\usepackage{placeins}
\usepackage{enumerate}

\usepackage{multirow, booktabs}

\usepackage{pifont}% http://ctan.org/pkg/pifont
\newcommand{\cmark}{\ding{51}}%
\newcommand{\xmark}{\ding{55}}%

\newcommand{\name}{CIOSL}

\DeclareMathOperator*{\argmin}{arg\,min}

\DeclareMathOperator*{\argmax}{arg\,max}

\usepackage{algorithm}
\usepackage{algorithmic}

\def\cvprPaperID{*****} % *** Enter the CVPR Paper ID here
\def\confName{CVPR}
\def\confYear{2022}

\begin{document}

\title{Class Incremental Online Streaming Learning}

\author{
Soumya Banerjee$^{1}$, \hspace{0.05cm} Vinay Kumar Verma$^{2}$, \hspace{0.05cm} Toufiq Parag$^3$, \hspace{0.05cm} Maneesh Singh$^3$, \hspace{0.05cm} Vinay P. Namboodiri$^{1,4}$ \\
$^1$IIT Kanpur, India\hspace{0.15cm}
$^2$Duke University, USA\hspace{0.15cm}
$^3$Verisk Analytics, NJ, USA\hspace{0.15cm}
$^4$University of Bath, UK\\
{\tt\small soumyab@cse.iitk.ac.in, vinaykumar.verma@duke.edu, toufiq.parag@verisk.com}\\ 
{\tt\small maneesh.singh@verisk.com, vpn22@bath.ac.uk}
}

\maketitle

%%%%%%%%% ABSTRACT
\begin{abstract}

  A wide variety of methods have been developed to enable lifelong learning in conventional deep neural networks. However, to succeed, these methods require a `batch' of samples to be available and visited multiple times during training. While this works well in a static setting, these methods continue to suffer in a more realistic situation where data arrives in \emph{online streaming manner}. We empirically demonstrate that the performance of current approaches degrades if the input is obtained as a stream of data with the following restrictions: $(i)$ each instance comes one at a time and can be seen only once, and $(ii)$ the input data violates the i.i.d assumption, i.e., there can be a class-based correlation. We propose a novel approach (CIOSL) for the class-incremental learning in an \emph{online streaming setting} to address these challenges. The proposed approach leverages implicit and explicit dual weight regularization and experience replay. The implicit regularization is leveraged via the knowledge distillation, while the explicit regularization incorporates a novel approach for parameter regularization by learning the joint distribution of the buffer replay and the current sample. Also, we propose an efficient online memory replay and replacement buffer strategy that significantly boosts the model's performance. Extensive experiments and ablation on challenging datasets show the efficacy of the proposed method.

\end{abstract}

%%%%%%%%% BODY TEXT
\begin{figure*}[!htbp]
	%\frenchspacing
    \centering
    
    \includegraphics[width=14cm, height=2.1cm]{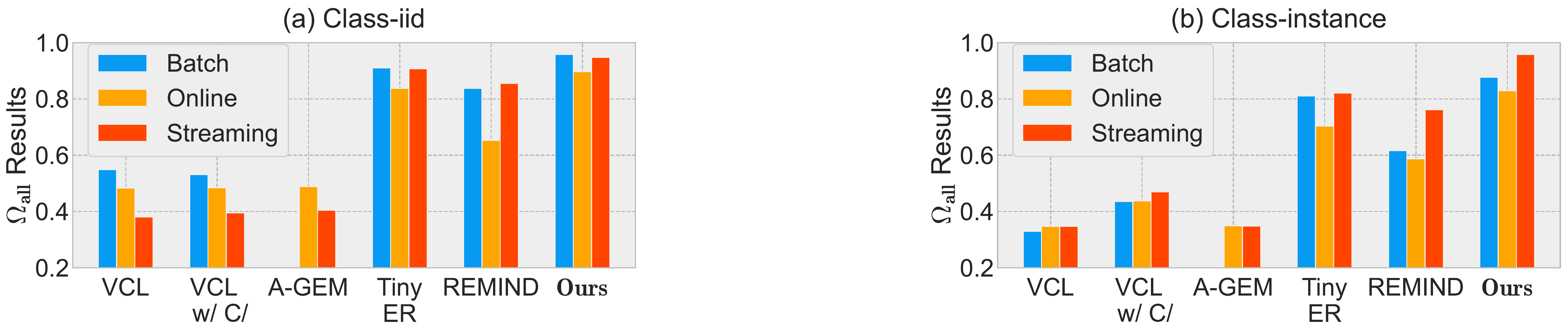}

    \caption{$\boldsymbol{\Omega}_{\text{all}}$ results for \emph{`batch'}, \emph{`online'} and \emph{`streaming'}  versions of baselines on iCubWorld 1.0 on $(i)$ Class-iid, and $(ii)$ Class-instance ordering. An empty plot in A-GEM indicates, we were unable to conduct experiment due to compatibility issues.} 
    \label{fig:batch_vs_online_vs_streaming}
    
\end{figure*}

\section{Introduction}\label{introduction}

% \frenchspacing
% \textcolor{red}{Consider that we are interested in deep learning networks that are trained continually, the best way to ensure that we do so effectively is by being able to learn from each sample being observed at a time. This is challenging as the samples could be observed in any order. Moreover, classical multiple epoch based training approaches would not be meaningful as it would lead to severe overfitting. We also want an ability to be able to use the model-on-the-fly without waiting for the termination of training. This implies a single pass update requirement for such a system. We show that through the present work we are able to propose a novel Bayesian class-incremental online streaming approach that can solve the problem effectively.}

In this paper, we aim to achieve true continual learning by solving an extreme and restrictive form of lifelong learning. Predominantly, the more popular and successful methods in continual learning operate in \emph{incremental batch learning} (IBL) scenarios~\cite{rusu2016progressive, shin2017continual, kirkpatrick2017overcoming, wu2018memory, aljundi2018memory}. In IBL, we assume that current task data is available in batches during training, and the learner visits them sequentially multiple times. However, these methods are ill-suited in a continuously changing environment, where the learner needs to quickly adapt to the newly available data without catastrophic forgetting~\cite{french1999catastrophic} in an \emph{online manner}.

The ability to continually learn effectively from streaming data with no catastrophic forgetting is a challenge that has not received widespread attention~\cite{hayes2019memory}. However, its utility is apparent, as it enables the practical deployment of autonomous AI agents. In the real world, an autonomous agent continuously encounters new examples of a novel or previously observed classes. To adapt to these new examples, it is often required to train the AI agent with these new instances immediately without disrupting its service. It is infeasible to wait and aggregate a `batch' of new samples for learning on the new data. Moreover, the ability to learn online in a \emph{streaming setting} is not only practical, but a step towards enabling true lifelong learning on embodied AI agents in a dynamic, non-stationary environment.

% In this work, we are interested in learning continuously in \emph{online streaming setting}, where a learner needs to learn from a single sample at a time with no catastrophic forgetting in a \emph{single-pass}, i.e., the data can be visited only once so that it can be evaluated immediately without waiting for termination of the training. 

In this work, we are interested in learning continuously in \emph{online streaming setting}, where a learner needs to learn from a single sample at a time, with no catastrophic forgetting, in a \emph{single-pass}. The model can visit any part of data only once, and it can be evaluated any time without waiting for termination of training. In addition, we also aim to evaluate the model in \emph{class-incremental setting}, such that at test time, we consider the label space over all the observed classes so far. This is in contrast to task-incremental learning methods such as VCL \cite{nguyen2017variational}, which requires task labels to be specified during inference. While some recent works aim at \emph{online learning} and consider \emph{class-incremental learning} setting, these methods are not suitable in the \emph{streaming learning} setting. For instance, GDumb~\cite{prabhu2020gdumb} suggests that storing data points with a greedy-sampler and retraining before inference outperforms existing approaches. However, one does visit data points multiple times during retraining, which violates the \emph{single-pass} learning constraint.  AGEM~\cite{chaudhry2018efficient}, another online learning approach, is based on projected-gradient-descent, also suffers severe forgetting in \emph{streaming setting}. Finally, the recently proposed \emph{streaming learning} approach, REMIND~\cite{hayes2019remind} is limited in terms of requiring a significant amount of cached data. Our work addresses all these limitations in a principled manner with a limited amount of information. Figure~\ref{fig:batch_vs_online_vs_streaming} compares the proposed method with the recent strong baselines. We observe that \emph{Our-model} outperforms the baselines by a significant margin in \emph{all three different lifelong learning} settings. It implies that a learning method in the most restrictive setting can be thought of as a universal method for all lifelong learning settings with the widest possible applicability.

We propose a novel method, \textbf{\emph{`Class Incremental Online Streaming Learning'} ({\name})}, which enables lifelong learning in \emph{online streaming setting} by leveraging tiny episodic memory replay and regularization. It regularises the model from two sources. One focuses on regularizing the model parameters explicitly in an online Bayesian framework, while the other regularises implicitly by enforcing the updated model to produce similar outputs for previous models on the past observed data. Our approach is jointly trained with buffer replay and current task samples by incorporating the likelihood of the replay and current samples. As a result, we do not need explicit finetuning with buffer samples. We propose a novel online loss-aware strategy for buffer replacement and coreset sample selection that significantly boosts the model's performance. Our approach only requires a small subset of samples for memory replay, and we do not use full buffer replay like past approaches~\cite{hayes2019memory}. 
Our experimental results on four benchmark datasets demonstrate the effectiveness as well as superiority of the proposed method to circumvent catastrophic forgetting over state-of-the-art baselines, and the extensive ablations validate the components of our method.

% As a result, it is computationally highly efficient. We perform experiments on four benchmark datasets and compared with state-of-the-art baselines. Our experimental results show the effectiveness of the proposed method and robustness to data orderings that can induce catastrophic forgetting. We also conduct extensive ablations to validate the components of our method.

Our contributions can be summarised as follows: $(i)$ we propose a novel dual regularization framework \textbf{({\name})}, comprising an online Bayesian framework as well as a functional regularizer, to overcome catastrophic forgetting in challenging \emph{online streaming learning} scenario, $(ii)$ we propose a novel online loss-aware buffer replacement and sampling strategy which significantly boosts the model's performance, $(iii)$ we empirically show that selecting a subset of samples from memory and computing the joint likelihood with the current sample is highly efficient, and enough to avoid explicit finetuning, and $(iv)$ we experimentally show that our method significantly outperforms the state-of-the-art baselines.

% We conduct extensive experiments and ablations on the challenging datasets to validate the efficacy of the proposed method. Our contributions can be summarized as follows:

% $\bullet$ We propose a dual regularization framework to mitigate forgetting in challenging \emph{online streaming learning scenarios}.

% $\bullet$

% We perform experiments on four datasets and compared with state-of-the-arts methods and baselines. Our experimental results shows the effectiveness of the proposed method and robustness to data orderings that can induce catastrophic forgetting. We also conduct extensive experiments to validate the components of our method. Our contributions can be summarized as follows:

\section{Problem Formulation}

% \emph{Online streaming learning} (OSL) or \emph{streaming learning} is the extreme case of online learning, where the data arrives sequentially one at a time, and the learner needs to learn in a \emph{`single-pass'}, i.e., the data can be visited only once. Let us consider that we have a dataset $\mathcal{D}$ with $T$ task sequences, i.e. $\mathcal{D} = {\{ \mathcal{D}_{t} \}}_{t = 1}^{T}$, where $\mathcal{D}_{t} = {\{ (\boldsymbol{x}_{j}, y_{j} ) \}}_{j = 1}^{N_{t}}$. In \emph{streaming learning}, it is assumed that $N_{t} = 1$ for all $t$, unlike the \emph{incremental batch learning} (IBL) and \emph{online learning} approaches where it is assumed $N_{t} >> 1$ for all $t$. In addition, the model needs to adapt to any part of the dataset without looping over multiple times, i.e., the model needs to learn in a \emph{`single-pass'}, contrary to the \emph{incremental batch learning}~\cite{rusu2016progressive,shin2017continual,kirkpatrick2017overcoming,aljundi2018memory} and \emph{online learning}~\cite{chaudhry2018efficient, lopez2017gradient, prabhu2020gdumb} approaches which allow visiting data-batches $\mathcal{D}_{t}$ sequentially for multiple times and fine-tuning the network parameters with the samples stored in memory respectively. Furthermore, the data coming in \emph{streaming setting} can be temporally contiguous, i.e., there could be a class based correlation, and the memory usage must be minimal.

\emph{Online streaming learning} (OSL) or \emph{streaming learning} is the extreme case of online learning, where the data arrives sequentially one at a time, and the model needs to learn in a \emph{`single-pass'}. Let us consider that we have a dataset $\mathcal{D}$ with $T$ task sequences, i.e. $\mathcal{D} = {\{ \mathcal{D}_{t} \}}_{t = 1}^{T}$, where $\mathcal{D}_{t} = {\{ (\boldsymbol{x}_{j}, y_{j} ) \}}_{j = 1}^{N_{t}}$. In \emph{streaming learning}, it is assumed that $N_{t} = 1$ for all $t$, unlike \emph{incremental batch learning} (IBL) and \emph{online learning} approaches, where it is assumed $N_{t} \gg 1$ for all $t$. In addition, the model cannot loop over any part of the dataset, i.e., \emph{single-pass} learning, and it can be evaluated immediately rather waiting for termination of training. This is in contrast to the \emph{incremental batch learning}~\cite{kirkpatrick2017overcoming,aljundi2018memory} and \emph{online learning}~\cite{prabhu2020gdumb} approaches which allow visiting data-batches $\mathcal{D}_{t}$ sequentially for multiple times and fine-tuning the network parameters with the buffer samples before inference, respectively. Furthermore, the data coming in \emph{streaming setting} can be temporally contiguous, i.e., there could be a class based correlation, and the memory usage must be minimal. Finally, the model is evaluated in \emph{class incremental setting}, such that, at test time, the label space is considered over all the classes observed so far.

\begin{figure}[t]
    \centering
    \includegraphics[width=8cm, height=1.8cm]{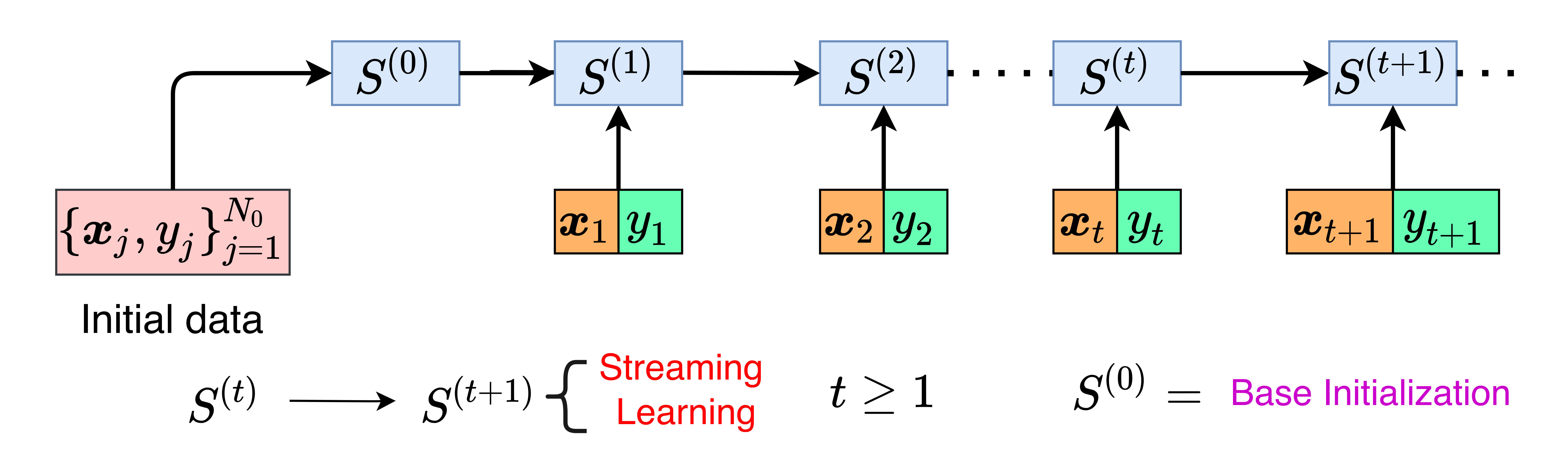}
    % \includegraphics[scale=0.43]{streaming_learning_framework.png}
    % \caption{\textbf{Proposed streaming learning framework.} $h^{(t)}$ represents the evolving state of the model after learning the $t$-th example, i.e., $\left( \boldsymbol{x}_{t}, y_{t} \right)$. In base initialization, i.e., state $h^{(0)}$, the model is trained offline with initial dataset. Then, in each incremental step, it is trained with one sample at a time.} 
    \caption{\textbf{Proposed streaming learning framework.} $S^{(t)}$ represents the evolving state of the model after learning the $t$-th example, i.e., $\mathcal{D}_{t} = \{ \boldsymbol{d}_{t} \} = \{(\boldsymbol{x}_{t}, y_{t})\}$.} 
    \label{fig:proposed_learning_framework}
\end{figure}

\section{Online Streaming Learning Framework}\label{online_streaming_learning_framework}

In this section, we introduce a \emph{`class incremental online-streaming learning'} framework ({\name}) which trains a convolutional neural network (CNN) in \emph{streaming setting}, as depicted in Figure~\ref{fig:proposed_learning_framework}. Formally, a limited number of samples are used to train the model offline during base initialization, $S^{(0)}$. Then in each incremental step, $\forall t > 0, S^{(t)}$,  it observes a new example $\left( \boldsymbol{x}_{t}, y_{t} \right)$, and the parameters are updated with a single step posterior computation. To avoid catastrophic forgetting, we use dual regularization by proposing implicit and explicit regularization over the model parameters. Section~\ref{learning_in_online_streaming_scenario} discusses the detailed regularization model. We also propose a novel online samples selection strategy for replay and buffer replacement which significantly improves model performance. We use a tiny episodic memory, and select only few informative past samples for replay \emph{instead of replaying the whole buffer}, Section~\ref{informative_past_sample_selection_for_memory_replay} and ~\ref{memory_buffer_replacement_policy} are dedicated to the detailed discussion about the buffer policy.

% In this work, we aim to train a convolutional neural network (CNN) in a streaming setting without suffering from catastrophic interference.

Formally, we separate the CNN into two neural networks: $(i)$ \emph{non-plastic} feature extractor $G(\cdot)$ consisting the first few layers of the CNN, and $(ii)$ \emph{plastic} neural network $F(\cdot)$ consisting the final layers of the CNN. For a given input image $\boldsymbol{x}$, the predicted class label is computed as: $y = F(G(\boldsymbol{x}))$. We initialize the parameters of the feature extractor $G(\cdot)$ and keep it frozen throughout online streaming setting. We use a \emph{Bayesian-neural-network} (BNN)~\cite{neal2012bayesian} as the \emph{plastic} network $F(\cdot)$, and optimize its parameters with sequentially coming data in \emph{streaming setting}.

\subsection{Learning in Online Streaming Scenario}\label{learning_in_online_streaming_scenario}

\begin{figure}[t]
  \centering
  \includegraphics[width=6.8cm, height=2.7cm]{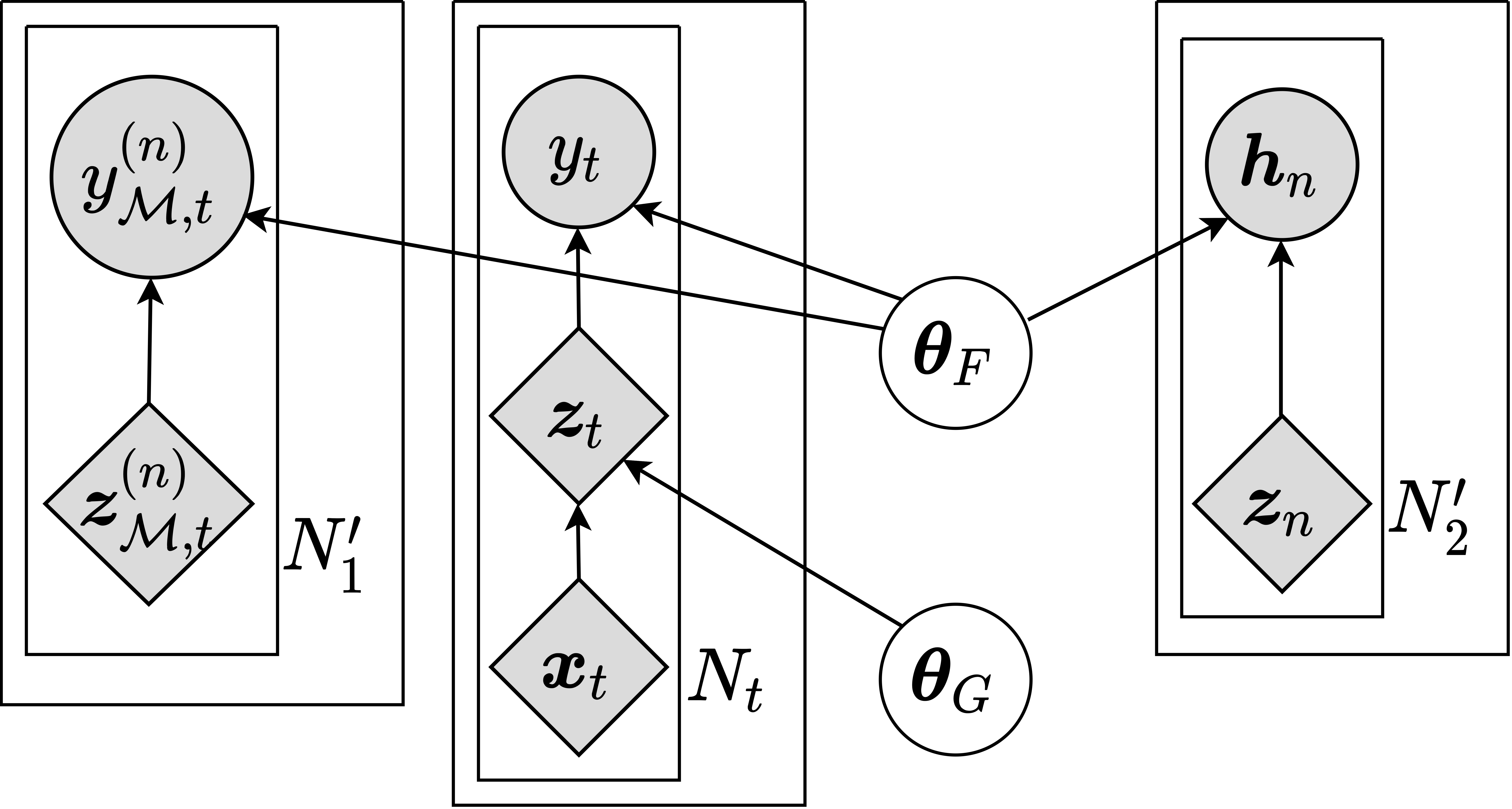}
  \caption{Graphical model of the proposed neural network training in \emph{`streaming-setting'}, as discussed in Section~\ref{learning_in_online_streaming_scenario}.} 
  \label{fig:proposed_streaming_learning_model}
  % \vspace{-4mm}
\end{figure}

Online updating naturally emerges from the Bayes' rule; given the posterior $p(\boldsymbol{\theta} | \mathcal{D}_{1:{t - 1}})$, whenever a new data $\mathcal{D}_{t} = \{ \boldsymbol{d}_{t} \} = \{(\boldsymbol{x}_{t}, y_{t})\}$ comes in, we can compute a new posterior $p(\boldsymbol{\theta} | \mathcal{D}_{1:t})$ by combining the previous posterior and the new data-likelihood, i.e., $p(\boldsymbol{\theta} | \mathcal{D}_{1:t}) \propto p(\mathcal{D}_{t} | \boldsymbol{\theta}) \; p(\boldsymbol{\theta} | \mathcal{D}_{1:{t - 1}})$, where the old posterior is treated as the prior. However, for any complex model, exact Bayesian inference is not tractable, and an approximation is needed. A Bayesian neural network commonly approximates the posterior with a variational posterior $q(\boldsymbol{\theta})$ by maximizing the evidence-lower-bound (ELBO)~\cite{blundell2015weight}. However, optimizing the ELBO to approximate the posterior $p(\boldsymbol{\theta} | \mathcal{D}_{1:t})$ only with data $\mathcal{D}_{t}$ can fail in streaming setting~\cite{ghosh2018structured}. Furthermore, we evaluate the model's performance in \emph{class incremental setting}, such that, at test time the label space is considered over all the classes have been observed till the time instance $t$. Moreover, training only with $\mathcal{D}_{t}$ makes the model biased towards the new data or task and parameter regularization is not sufficient to overcome forgetting. To overcome these limitations, we include a \emph{`fixed-sized'} tiny episodic memory for storing a small representative subset $(\leq 5\%)$ of all the observed samples. Instead of storing the raw input $\boldsymbol{x}$, we store the embedding $\boldsymbol{z} = G(\boldsymbol{x})$, where $\boldsymbol{z}\in \mathbb{R}^d$. Storing the embeddings save a significant amount of memory, and also saves the model from doing a forward pass through the convolutional layers, making the training procedure computationally highly efficient.

During training, we select a subset of samples from memory \emph{instead of replaying the whole buffer}, and replay them with the new data $\mathcal{D}_{t} = \{ \boldsymbol{d}_{t} \} = \{(\boldsymbol{x}_{t}, y_{t})\}$.
Therefore, the new posterior computation can be written as follows: $p(\boldsymbol{\theta} | \mathcal{D}_{1:t}) \propto p(\mathcal{D}_{t} | \boldsymbol{\theta}) p(\mathcal{D}_{\mathcal{M}, t} | \boldsymbol{\theta}) p(\boldsymbol{\theta} | \mathcal{D}_{1: {t - 1}})$, which we approximate with a variational posterior $q_{t}(\boldsymbol{\theta})$ as follows: 
\begin{equation}\label{eq_2}
% \small
\mbox{\fontsize{8.4pt}{10.2pt}\selectfont\( %
    \mathcal{L}^{1}_{t}(\boldsymbol{\theta}) = \argmin_{q \epsilon \mathcal{Q}} \text{KL}\left( q_{t}(\boldsymbol{\theta}) \left|\right| \frac{1}{Z_{t}} q_{t - 1}(\boldsymbol{\theta}) p(\mathcal{D}_{t} | \boldsymbol{\theta}) p(\mathcal{D}_{\mathcal{M}, t} | \boldsymbol{\theta})  \right)
\)} %
\end{equation}

\noindent where $\mathcal{D}_{\mathcal{M}, t} \subset \mathcal{M}$ represents the subset of samples selected from the memory $\mathcal{M}$ for replaying at time $t$, and $\mathcal{D}_{t}$ represents the new data arriving at time $t$. Note that Eq. (\ref{eq_2}) is significantly different from VCL~\cite{nguyen2017variational}, where they assume \emph{multi-task/task-incremental learning setting} with separate \emph{head networks}, and incorporates the coreset samples only during explicit finetuning before inference.

% Note that Eq. (\ref{eq_2}) is significantly different from the VCL where they don't have a way to incorporate the replay samples and do the explicit finetuning during test, which increases the model complexity and inefficiently utilize the replay samples.

The above minimization is equivalent to maximization of the evidence-lower-bound (ELBO):
\begin{align}\label{eq_3}
\small
% \mbox{\fontsize{8pt}{9.6pt}\selectfont\( %
\mathcal{L}^{1}_{t}(\boldsymbol{\theta}) &= \begin{multlined}[t]
\mathbb{E}_{\boldsymbol{\theta} \sim q_{t}(\boldsymbol{\theta})} \left[ \log{p(y_{t} | \boldsymbol{\theta}, G(\boldsymbol{x}_{t}))} \right] \\ 
+ \sum_{n = 1}^{N_{1}^{\prime}} \mathbb{E}_{\boldsymbol{\theta} \sim q_{t}(\boldsymbol{\theta})} \left[ \log{p(y_{\mathcal{M}, t}^{(n)} | \boldsymbol{\theta}, \boldsymbol{z}_{\mathcal{M}, t}^{(n)})} \right] \\
- \lambda_{1} \cdot \text{KL}\left(q_{t}(\boldsymbol{\theta}) \left|\right| q_{t - 1}(\boldsymbol{\theta}) \right)
\end{multlined}
% \)} %
\end{align}

\noindent where $\mathcal{D}_{t} = \{ \boldsymbol{d}_{t} \} = \{ ( \boldsymbol{x}_{t}, y_{t} ) \}$, $\mathcal{D}_{\mathcal{M}, t} = \{ \boldsymbol{d}^{(n)}_{\mathcal{M}, t} \}^{N_{1}^{\prime}}_{n = 1} = \{ ( \boldsymbol{z}_{\mathcal{M}, t}^{(n)}, y_{\mathcal{M}, t}^{(n)} ) \}^{N_{1}^{\prime}}_{n = 1}$, $| \mathcal{D}_{\mathcal{M}, t} | = N_{1}^{\prime} \ll | \mathcal{M} |$, $\mathcal{D}_{\mathcal{M}, t} \subset \mathcal{M}$, and $\lambda_{1}$ is a hyper-parameter. 
%\textcolor{red}{The detailed derivation is given in the appendix.}

While the KL-divergence minimization (in Eq. (\ref{eq_3})) between prior and the posterior ensures minimal changes in the network parameters, initialization of the prior with the posterior at each time can introduce information loss in the network for a longer sequence of streaming data. We overcome such limitation with a functional/implicit regularizer, which encourages the network to mimic the output responses as produced in the past for the observed samples. Specifically, we minimize the KL-divergence between the class-probability scores obtained in past and current time $t$:
\begin{equation}\label{eq_4}
\small
    \sum^{t - 1}_{n = 1} \mathbb{E}_{\boldsymbol{\theta} \sim q_{t}(\boldsymbol{\theta})} \left[ \text{KL} \left( \text{softmax}(\boldsymbol{h}_{n}) \; \left|\right| \; F_{\boldsymbol{\theta}}(G(\boldsymbol{x}_{n})) \right) \right]
\end{equation}

\noindent where $\boldsymbol{x}_{n}$ and $\boldsymbol{h}_{n}$ represents input examples and the logits obtained while training on $\mathcal{D}_{n}$ respectively.

However, the objective in Eq. (\ref{eq_4}) requires the availability of the embeddings and the corresponding logits for all the past observed data. Since storing all the past data is not feasible, we only store the logits for all samples in memory $\mathcal{M}$. During training, we uniformly select $N_{2}^{\prime}$ samples along with their logits and optimize the following objective:
\begin{equation}\label{eq_5}
\small
   \mathcal{L}^{2}_{t}(\boldsymbol{\theta}) = \sum^{N_{2}^{\prime}}_{n = 1} \mathbb{E}_{\boldsymbol{\theta} \sim q_{t}(\boldsymbol{\theta})} \left[ \text{KL} \left( \text{softmax}(\boldsymbol{h}_{n}) \; \left|\right| \; F_{\boldsymbol{\theta}}(\boldsymbol{z}_{n}) \right) \right]
\end{equation}

\noindent where (i) $\boldsymbol{z}_{n}$ and $\boldsymbol{h}_{n}$ represents feature-map and corresponding logits, and (ii) $N_{2}^{\prime} \ll \left| \mathcal{M} \right|$.

Under the mild assumptions of knowledge distillation~\cite{hinton2015distilling}, the optimization in Eq. (\ref{eq_5}) is equivalent to minimization of the Euclidean distance between the logits, and the optimization objective can be written as:
\begin{equation}\label{eq_6}
\small
    \mathcal{L}^{2}_{t}(\boldsymbol{\theta}) = \lambda_{2} \cdot \sum^{N_{2}^{\prime}}_{n = 1} \mathbb{E}_{\boldsymbol{\theta} \sim q_{t}(\boldsymbol{\theta})} \left[ \left|\left| \boldsymbol{h}_{n} - f_{\boldsymbol{\theta}}(\boldsymbol{z}_{n}) \right|\right|^{2}_{2} \right]
\end{equation}
\noindent where $(i)$ $f(\cdot)$ represents the plastic network $F(\cdot)$ without the softmax activation, and $(ii)$ $\lambda_{2}$ is a hyper-parameter.

Training the plastic network (BNN) $F(\cdot)$ requires specification of $q(\boldsymbol{\theta})$ and, in this work, we model $\boldsymbol{\theta}$ by stacking up the parameters (weights \& biases) of the network $F(\cdot)$. We use a Gaussian mean-field posterior $q_{t}(\boldsymbol{\theta})$ for the network parameters, and choose the prior distribution, i.e., $q_{0}(\boldsymbol{\theta}) = p(\boldsymbol{\theta})$, as multivariate Gaussian distribution. We train the network $F(\cdot)$ by maximizing the ELBO in Eq. (\ref{eq_3}) and minimizing the Euclidean distance in Eq. (\ref{eq_6}). For memory replay in Eq. (\ref{eq_3}), we select samples using strategies mentioned in Section~\ref{informative_past_sample_selection_for_memory_replay}, and we use \emph{uniform sample} selection strategy to select samples from memory to be used in Eq. (\ref{eq_6}).

% \section{Memory Replay Policy}

\subsection{Informative Past Sample Selection For Replay}\label{informative_past_sample_selection_for_memory_replay}

% We consider the following strategies:

\textbf{Uniform Sampling (Uni).} In this approach, samples are selected uniformly random from the memory. %If we have the buffer of size $K$ then each samples are selected with probability $1/K$.

\textbf{Uncertainty-Aware Positive-Negative Sampling (UAPN).} UAPN selects ${N_{1}^{\prime}}/{2}$ samples with the highest uncertainty scores (negative samples) and ${N_{1}^{\prime}}/{2}$ samples with the lowest uncertainty scores (positive samples). Empirically, we observe that this sample selection strategy results in the best performance. We measure the predictive uncertainty~\cite{chai2018uncertainty} for an input $\boldsymbol{z}$ with BNN $F(\cdot)$ as follows: 
\begin{equation}\label{eq_7}
\mbox{\fontsize{8.5pt}{10.2pt}\selectfont\( %
    \Phi (\boldsymbol{z}) \approx - \begin{multlined}[t]
    \sum_{c} \left( \frac{\sum_{k} p\left(\hat{y}=c|{\boldsymbol{z}},{\boldsymbol{\theta}}^{k}\right)}{k}  \right) 
  \log \left( \frac{\sum_{k} p\left(\hat{y}=c|{\boldsymbol{z}},{\boldsymbol{\theta}}^{k}\right)}{k}  \right)
    \end{multlined}
\)} %
\end{equation}
\noindent where $p(\hat{y}=c|{\boldsymbol{z}},{\boldsymbol{\theta}}^{k})$ is the predicted softmax output for class $c$ using the $k$-th sample of weights $\boldsymbol{\theta}^{k}$ from $q(\boldsymbol{\theta})$. We use $k = 5$ samples for uncertainty estimation.

\textbf{Loss-Aware Positive-Negative Sampling (LAPN).} LAPN selects ${N_{1}^{\prime}}/{2}$ samples with the highest loss-values (negative-samples), and ${N_{1}^{\prime}}/{2}$ samples with the lowest loss-values (positive-samples). Empirically we observe that the combination of most and least certain samples shows a significant performance boost since one ensures quality while the other ensures diversity for the memory replay.

\subsection{Memory Buffer Replacement Policy}\label{memory_buffer_replacement_policy}
For memory replay, we use a \emph{`fixed-sized'} tiny episodic memory. However, in a lifelong learning setting, data may come indefinitely, implying that the capacity of the replay buffer will be quickly exhausted. To combat such an issue, we employ a replay buffer replacement policy which replaces a previously stored sample with a new instance if the buffer is full. Otherwise, the new instance is stored.

\textbf{Loss-Aware Weighted Class Balancing Replacement (LAWCBR).} In this approach, whenever a new sample comes in and the buffer is full, we remove a sample from the class with maximum number of samples present in the buffer $(y_{r} = \argmax \text{ClassCount}(\mathcal{M}))$. However, instead of removing an example uniformly, we weigh each sample of the majority class inversely w.r.t their loss $(w^{y_{r}}_{i} \propto \frac{1}{l^{y_{r}}_{i}})$ and use these weights as the replacement probability; the lesser the loss, the more likely to be removed.

\textbf{Loss-Aware Weighted Random Replacement With A Reservoir (LAWRRR).}   In this approach, we propose a novel variant of reservoir sampling~\cite{vitter1985random} to replace an existing sample with the new sample when the buffer is full. We weigh each stored sample inversely w.r.t the loss $(w_{i} \propto \frac{1}{l_{i}})$, and proportionally to the total number of examples of that class in which the sample belongs present in the buffer $(w_{i} \propto \text{ClassCount}(\mathcal{M}, y_{i}))$. Whenever a new example satisfies the replacement condition of reservoir sampling, we combine these two scores and use that as the replacement probability; the higher the weight, the more likely to be replaced.

\subsection{Making Sampling Strategies Online}\label{making_sampling_strategies_online}
Loss-aware sampling strategies require computing the loss-values of each stored sample in each incremental step, resulting in a computationally expensive learning process. To overcome this ssue, we propose the following online update policy of the loss-values: $(i)$ for each sample stored in memory, we keep the corresponding loss-value, $(ii)$ whenever a sample is selected for memory reply, we replace the previously computed loss-value with the new loss-value at time $t$; furthermore, we update the past logits with the newly computed logits, as changes in the loss-value reflect changes in the logits, too. To accommodate the online updating of loss-values, we propose to keep the loss-value in memory for each stored sample; however, since it is just a scalar value, the storage cost is negligible. To make uncertainty-aware sampling online, we store the uncertainty scores in memory and update them in a similar manner; the storage cost remains negligible, as it is another scalar value.

\subsection{Feature Extractor}
\label{feature_extractor}

In this work, we separate the representation learning, i.e., learning the feature extractor $G(\cdot)$, and the classifier learning, i.e., learning the plastic network $F(\cdot)$. Similar to several existing continual learning approaches~\cite{kemker2017fearnet, hayes2019memory, xiang2019incremental, hayes2019remind}, we initialize the feature extractor $G(\cdot)$ with the weights learned through supervised visual representation learning~\cite{krizhevsky2012imagenet}, and keep them fixed throughout \emph{streaming learning}. The motivation to use a pre-trained feature extractor is that the features learned by the first few layers of the neural networks are highly transferable and not specific to any particular task or dataset and can be applied to several different task(s) or dataset(s)~\cite{yosinski2014transferable}. Moreover, it is hard to learn generalized visual features, which can be used across all the classes~\cite{zhu2021prototype} with having access to only a single example at each time instance.

In our experiments, for all the baselines along with {\name}, we use Mobilenet-V2~\cite{sandler2018mobilenetv2} pre-trained on ImageNet~\cite{russakovsky2015imagenet} as the base architecture for the visual feature extractor. It consists of a convolutional base and a classifier network. We remove the classifier network and use the convolutional base as the feature extractor $G(\cdot)$ to obtain embedding that is fed to the plastic network BNN $F(\cdot)$. For details on the plastic network used for other baselines, refer to Section~\ref{implementation_details}.

\section{Related Work}\label{related_work}

% Existing continual learning approaches can be broadly classified into~\cite{parisi2019continual}: (i) parameter-isolation-based approaches, (ii) regularization-based approaches, and (iii) rehearsal-based approaches.

Parameter-isolation-based approaches train different subsets of model parameters on sequential tasks. PNN~\cite{rusu2016progressive}, DEN~\cite{yoon2017lifelong} expand the network to accommodate the new task. PathNet~\cite{fernando2017pathnet}, PackNet~\cite{mallya2018packnet}, Piggyback~\cite{mallya2018piggyback}, and HAT~\cite{serra2018overcoming} train different subsets of network parameters on each task.

% identify the parts of the model used previously and mask them when training the model on the new task.

Regularization-based approaches use an extra regularization term in the loss function to enable continual learning. LWF~\cite{li2017learning} uses knowledge distillation~\cite{hinton2015distilling} loss to prevent catastrophic forgetting. EWC~\cite{kirkpatrick2017overcoming}, IMM~\cite{lee2017overcoming}, and MAS~\cite{aljundi2018memory} regularize by penalizing changes to the important weights of the network.

Rehearsal-based approaches replay a subset of past training data during sequential learning. iCaRL~\cite{rebuffi2017icarl}, SER~\cite{isele2018selective}, and TinyER~\cite{chaudhry2019continual} use memory replay when training on a new task. DER~\cite{buzzega2020dark} uses knowledge distillation and memory replay while learning a new task. DGR~\cite{shin2017continual}, MeRGAN~\cite{wu2018memory}, and CloGAN~\cite{rios2018closed} retain the past task(s) distribution with a generative model and replay the synthetic samples during incremental learning. Our approach also leverages memory replay from a tiny episodic memory; however, we store the feature-maps instead of raw inputs.

% \textcolor{red}{Variational Continual Learning (VCL)~\cite{nguyen2017variational} leverages Bayesian inference to mitigate catastrophic forgetting. However, the approach naively adapted performs poorly for the streaming learning setting. Additional refinements by using sample selection using coreset (it would also require the task-id during inference) while improving the approach would violate the single pass update setting. Moreover it still does not outperform our approach even with fine-tuning. Our approach addresses all the above issues in a principled manner and alleviates catastrophic forgetting in \emph{online streaming setting}}

Variational Continual Learning (VCL)~\cite{nguyen2017variational} leverages Bayesian inference to mitigate catastrophic forgetting. However, the approach, when na\"ively adapted, performs poorly in the \emph{streaming learning setting}. Additionally, it also needs task-id during inference. Furthermore, the explicit finetuning with the buffer samples (coreset) before inference violates the \emph{single-pass} learning constraint. Moreover, it still does not outperform our approach even with the finetuning. More details are given in the appendix.

% \textcolor{red}{REMIND~\cite{hayes2019remind} is a recently proposed rehearsal-based lifelong learning approach, which combats catastrophic forgetting in \emph{online streaming setting}. While it follows a setting close to the one proposed, the model stores a large number of past examples compared to the other baselines; for example, iCaRL~\cite{rebuffi2017icarl} stores 20K past examples for the ImageNet experiment, whereas REMIND stores 1M past examples. It actually uses a lossy compression for storing the samples. Such techniques can be used in any continual learning approach.}

REMIND~\cite{hayes2019remind} is a recently proposed rehearsal-based lifelong learning approach, which combats catastrophic forgetting in \emph{online streaming setting}. While it follows a setting close to the one proposed, the model stores a large number of past examples compared to the other baselines; for example, iCaRL~\cite{rebuffi2017icarl} stores 10K past examples for the ImageNet experiment, whereas REMIND stores 1M past examples. Further, it actually uses a lossy compression to store past samples, which is merely an engineering technique, not an algorithmic improvement, and can be used by any continual learning approach. For more details, please refer to the appendix.

% REMIND~\cite{hayes2019remind} is a recently proposed rehearsal-based lifelong learning approach, which combats catastrophic forgetting in \emph{online streaming setting}. Learning involves compression of each new input using product quantization (PQ), reconstruction of the previously stored compressed representations, mixing the past reconstructed examples with the new input, and updating the parameters of the plastic layers of the network. While following the challenging setting, the model stores considerably a large number of past examples compared to the baselines; for example, iCaRL~\cite{rebuffi2017icarl} stores 20K past examples for the ImageNet experiment, whereas REMIND stores 1M past examples. Also, because of the lossy compression in PQ, REMIND losses a significant amount of information, and compression and decompression include extra overhead.

\section{Experiments}\label{experiments}

\subsection{Baselines And Compared Methods}\label{baselines_and_compared_methods}

The proposed approach follows the \emph{`online streaming setting'}; to the best of our knowledge, recent works ExStream~\cite{hayes2019memory} and REMIND~\cite{hayes2019remind} are the only methods that follow this setting. We compare our approach against these strong baselines. We also compare our model with $(i)$ a network trained with a sample one at a time (Fine-tuning/lower-bound) and $(ii)$ a network trained offline, assuming all the data is available (Offline/upper-bound). Also, for rigorous comparison, we choose recent popular `batch' and `online' continual learning methods, such as EWC~\cite{kirkpatrick2017overcoming}, MAS~\cite{aljundi2018memory}, VCL with/without coreset~\cite{nguyen2017variational}, Coreset Only~\cite{farquhar2018towards}, TinyER~\cite{chaudhry2019continual}, GDumb~\cite{prabhu2020gdumb} and A-Gem~\cite{chaudhry2018efficient}. For a fair comparison, we train all the methods in an online streaming setting, i.e., one sample at a time. `VCL w/ C/' and `Coreset only' are both trained in a streaming manner; however, the network is fine-tuned with the stored samples before inference. Also, GDumb stores samples in memory and fine-tunes the network with them before inference, while fine-tuning is prohibited in \emph{`streaming setting'}. Therefore, VCL and GDumb have an extra advantage compared to the true \emph{`streaming learning'} approaches. Still, {\name} outperforms these approaches by a significant margin.  
More details are given in the appendix.

\begin{table}[!htbp]
% \small
\scriptsize
\centering
% \caption{Categorization of the baseline approaches  depending on
% the underlying simplifying assumptions they impose. V.C.SL: Violates Constraints of Streaming Learning}
% \label{Tabel_baseline_categorization}

\begin{tabular}{ c  c  c  c  c  c } %\hline
  \toprule

  \textbf{Method} & \shortstack{\textbf{Learning} \\ \textbf{Type}} & \shortstack{\textbf{Fine-tunes}} & \shortstack{\textbf{V.C.SL}} & \textbf{Regularize}  & \textbf{Memory} \\
  
  \midrule
  
    EWC & Batch & \xmark & \xmark & \cmark & \xmark \\
    
    MAS & Batch & \xmark & \xmark & \cmark & \xmark \\
    
    VCL & Batch & \xmark & \xmark & \cmark & \xmark \\
    
    VCL w/ C/ & Batch & \textcolor{red}{\cmark} & \textcolor{red}{\cmark} & \cmark & \cmark \\
    
    Coreset & Online & \textcolor{red}{\cmark} & \textcolor{red}{\cmark} & \xmark & \cmark \\
    
    GDumb & Online & \textcolor{red}{\cmark} & \textcolor{red}{\cmark} & \xmark & \cmark \\
  
    TinyER & Online & \xmark & \xmark & \xmark & \cmark \\
    
    A-GEM & Online & \xmark & \xmark & \cmark & \cmark \\
    
    ExStream & Streaming & \xmark & \xmark & \xmark & \cmark \\
    
    % Deep-SLDA & Streaming & No & No \\
    
    REMIND & Streaming & \xmark & \xmark & \xmark & \cmark \\
    
    % \midrule
    
    \textbf{Ours} & Streaming & \xmark & \xmark & \cmark & \cmark \\

 \bottomrule
    % \hline
\end{tabular}

\caption{Categorization of the baseline approaches  depending on
the underlying simplifying assumptions they impose. V.C.SL: Violates Constraints of Streaming Learning}
\label{Tabel_baseline_categorization}

\end{table}

%%%%%%%%%%%%%%%%%%%%%%%%%%%%%%%%%%%%%%%%%%%%%%%%%%%%%%%%%%%%%%%%%
%% Main Results Table
%%%%%%%%%%%%%%%%%%%%%%%%%%%%%%%%%%%%%%%%%%%%%%%%%%%%%%%%%%%%%%%%%

\begin{table*}[!htbp]
  \scriptsize
  % \small
  \centering
%   \caption{$\boldsymbol{\Omega}_{\text{all}}$ results. For each experiment, the method with best performance in \emph{`streaming-setting'} is highlighted in \textbf{Bold}. The reported results are average over $10$ runs with different permutations of the data. Offline model is trained only once. ${\widehat{\text{Offline}} =  \frac{1}{T}{\sum_{t = 1}^{T}{\boldsymbol{\alpha}_{\text{offline}, t}}}}$, where $T$ is the total number of testing events. `-' indicates
%   experiments we were unable to run, because of compatibility issues. Methods in \textcolor{red}{Red} use fine-tuning before inference, which violates \emph{`single-pass'} learning constraint.}
%   \label{Table_results_main_paper}

  \scriptsize
  % \small
  \begin{tabular}{ c | c c c c | c c c c | c | c } %\hline
    \toprule
    \multirow{2}{*} {\textbf{Method}} & \multicolumn{4}{c}{\textbf{iid}} & \multicolumn{4}{c}{\textbf{Class-iid}} & \multicolumn{1}{c}{\textbf{Instance}} &  \multicolumn{1}{c}{\textbf{Class-instance}} \\ 
    
    \cmidrule{2-11}

    & \shortstack{\textbf{CIFAR10}} & \shortstack{\textbf{CIFAR100}} & \shortstack{\textbf{ImageNet100}} & \shortstack{\textbf{iCubWorld}} & \shortstack{\textbf{CIFAR10}} & \shortstack{\textbf{CIFAR100}} & \shortstack{\textbf{ImageNet100}} & \shortstack{\textbf{iCubWorld}} & \shortstack{\textbf{iCubWorld}} & \shortstack{\textbf{iCubWorld}} \\

    \midrule 
      
      Fine-Tune & 0.1175 & 0.0180 & 0.0127 & 0.1369   & 0.3447 & 0.1277 & 0.1223 & 0.3893  & 0.1307 & 0.3485 \\
      
      EWC & - & - & - & -   & 0.3446 & 0.1292 & 0.1225 & 0.3790  & -  & 0.3487  \\
      
      MAS & - & - & - & -   & 0.3470 & 0.1280 & 0.1234 & 0.3912 & - & 0.3486 \\
      
      VCL & - & - & - & -   & 0.3442 & 0.1273 & 0.1205 & 0.3806 & - & 0.3473  \\
      
      \textcolor{red}{VCL w/ C/} & - & - & - & -  &  0.3716 & 0.1414 & 0.1259 & 0.3948  & -  & 0.4705  \\
      
      \textcolor{red}{Coreset}  & - & - & - & -  &  0.3684 & 0.1432 & 0.1273 & 0.3994  & -  & 0.4669  \\
      
      \textcolor{red}{GDumb} & 0.8686 & 0.6067 & 0.8361 & 0.8993   & \textit{\textbf{0.9252}} & 0.7635 & \textit{\textbf{0.9197}} & \textit{\textbf{0.9660}} & 0.6715  & 0.7908 \\
      
      A-GEM & 0.1175 & 0.0182 & 0.0139 & 0.1311   & 0.3448 & 0.1290 & 0.1215 & 0.4047 & 0.1309  & 0.3489  \\
      
      TinyER & 0.9314 & 0.7588 & 0.9415 & 0.9590  & 0.8926 & 0.7402 & 0.8995 & 0.9069  & 0.8726  & 0.8215  \\
      
      ExStream & 0.8866 & 0.7845 & 0.9293 & 0.9235  & 0.8123 & 0.7176 & 0.8757 & 0.8820  & 0.8954  & 0.8727  \\
      
      % Deep-SLDA &  &  &  &   &  &  &  &  &  &   \\
      
      REMIND & 0.8910 & 0.6457 & 0.9088 & 0.9260  & 0.8832 & 0.6787 & 0.8803 & 0.8553 & 0.8157  & 0.7615 \\
      
      % \midrule
      
      \textbf{Ours} & \textbf{0.9579} & \textbf{0.8679} & \textbf{0.9640} & \textbf{0.9716}  & \textbf{0.8991} & \textbf{0.7724} & \textbf{0.9171} & \textbf{0.9480}  & \textbf{0.9580}  & \textbf{0.9585} \\
     
    \midrule
     
    Offline & 1.0000 & 1.0000 & 1.0000  & 1.0000 & 1.0000 & 1.0000 & 1.0000 & 1.0000 & 1.0000  & 1.0000 \\
     
  %   \midrule
     
    $\widehat{\text{Offline}}$ & 0.8509 & 0.6083 & 0.8520 & 0.7626 & 0.8972 & 0.7154 & 0.8953 & 0.8849 & 0.7646 & 0.8840  \\ 
     
   \bottomrule
   
      % \hline
  \end{tabular}

  \caption{$\boldsymbol{\Omega}_{\text{all}}$ results. For each experiment, the method with best performance in \emph{`streaming-setting'} is highlighted in \textbf{Bold}. The reported results are average over $10$ runs with different permutations of the data. Offline model is trained only once. ${\widehat{\text{Offline}} =  \frac{1}{T}{\sum_{t = 1}^{T}{\boldsymbol{\alpha}_{\text{offline}, t}}}}$, where $T$ is the total number of testing events. `-' indicates
  experiments we were unable to run, because of compatibility issues. Methods in \textcolor{red}{Red} use fine-tuning before inference, which violates \emph{`single-pass'} learning constraint.}
  \label{Table_results_main_paper}

  \end{table*}

%%%%%%%%%%%%%%%%%%%%%%%%%%%%%%%%%%%%%%%%%%%%%%%%%%%%%%%%%%%%%%%%%%

\subsection{Datasets, Data Orderings And Metrics}\label{datasets_dataOrderings_metrics}

\textbf{Datasets.} To evaluate the efficacy of the proposed model we perform extensive experiments on four standard datasets: CIFAR10, CIFAR100~\cite{krizhevsky2009learning}, ImageNet100, and iCubWorld-1.0~\cite{fanello2013icub}. CIFAR10 and CIFAR100 are standard classification datasets with 10 and 100 classes, respectively. ImageNet100 is a subset of ImageNet-1000 (ILSVRC-2012)~\cite{russakovsky2015imagenet} containing randomly chosen 100 classes, with each class containing 700-1300 training samples and 50 validation samples, which are used for testing. 
iCubWorld 1.0 is an object recognition dataset which contains the sequences of video frames, with each containing a single object. There are 10 classes, with each containing 3 different object instances with 200-201 images each. Overall, each class contains 600-602 samples for training and 200-201 samples for testing. Technically, iCubWorld 1.0 is ideal dataset for \emph{streaming learning}, as it requires learning from temporally ordered image sequences, i.e., \emph{non-i.i.d} images.

\textbf{Evaluation Over Different Data Orderings.} The proposed approach is robust to the various streaming learning setting; we evaluate the model's streaming learning ability with the following four~\cite{hayes2019memory, hayes2019remind} challenging data ordering schemes: $(i)$ \emph{`streaming iid'}: where the data-stream is organized by the randomly shuffled samples from the dataset, $(ii)$ \emph{`streaming class iid`}: where the data-stream is organized by the samples from one or more classes, these samples are shuffled randomly, $(iii)$ \emph{`streaming instance'}: where the data-stream is organized by temporally ordered samples from different object instances, and $(iv)$ \emph{`streaming class instance'}: where the data-stream is organized by the samples from different classes, the samples within a class are temporally ordered based on different object instances. Only iCubWorld dataset contains the temporal ordering therefore \emph{`streaming instance'}, and \emph{`streaming class instance'} setting evaluated only on the iCubWorld dataset.
Please refer to the appendix for more details.

\textbf{Metrics.} For evaluating the performance of the streaming learner, we use $\boldsymbol{\Omega_{\text{all}}}$ metric, similar to \cite{kemker2018measuring, hayes2019memory, hayes2019remind}, where $\boldsymbol{\Omega}_{\text{all}}$ represents normalized incremental learning performance with respect to an offline learner: ${\boldsymbol{\Omega}_{\text{all}} = \frac{1}{T} \sum_{t = 1}^{T} \frac{\boldsymbol{\alpha}_{t}}{\boldsymbol{\alpha}_{\text{offline}, t}}}$, where $T$ is the total number of testing events, $\boldsymbol{\alpha}_{t}$ is the performance of the incremental learner at time $t$, and $\boldsymbol{\alpha}_{\text{offline}, t}$ is the performance of a traditional offline model at time $t$.

\subsection{Implementation Details}\label{implementation_details}

% We use Mobilenet-V2 \cite{sandler2018mobilenetv2} pre-trained on ImageNet \cite{russakovsky2015imagenet} available in PyTorch \cite{paszke2019pytorch} TorchVision package as the feature extractor $G(\cdot)$ to obtain embeddings from the raw pixels; we keep it frozen throughout the streaming learning.

In all the experiments, models are trained with one sample at a time. For a fair comparison, a similar network structure is used throughout all the models. For all methods, we use fully connected single-head networks with two hidden layers as the plastic network $F(\cdot)$, where each layer contains $256$ nodes with ReLU activations; for `VCL w/ or w/o Coreset', `Coreset only' and \emph{`{\name}'}, $F(\cdot)$ is a BNN, whereas for all other methods $F(\cdot)$ is a deterministic network. For a fair comparison, we store the same number of past examples for all replay-based approaches. For REMIND, we compress and store the feature-maps with Faiss \cite{johnson2019billion} product quantization (PQ) implementation with $s = 32$ sub-vectors and codebook size $c = 256$. 

We store the feature-map in memory for all the other methods, including our approach {\name}. In addition, {\name} also store the corresponding logits, loss-values, and uncertainty-scores. The capacity of our replay buffer is mentioned in Table~\ref{Tabel_buffer_capacity}. For memory-replay, we use \emph{`uncertainty-aware positive-negative'} sampling strategy (discussed in Sec:\ref{informative_past_sample_selection_for_memory_replay}) throughout all data-orderings, except for \emph{`streaming-i.i.d'} ordering, we use \emph{`uniform'} sampling. We use \emph{`loss-aware weighted random replacement with a reservoir'} sampling strategy as memory replacement policy for all the experiments. We store the same number of past examples in memory across all methods. 
%ExStream stores 6 prototypes for each class throughout all experiments. 
For memory-replay, we use $N_{1}^{\prime} = 16$ past samples throughout all experiments across {\name}, AGEM, TinyER, ExStream and REMIND. For knowledge-distillation, {\name} use $N_{2}^{\prime} = 16$ samples at any time step $t$. We set the hyper-parameter $\lambda_{1}= 1$ and $\lambda_{2} = 0.3$ across all experiments. For EWC, we set hyper-parameter $\lambda = 500$, and for MAS, we set hyper-parameter $\lambda = 1$. We repeated each experiments for 10 times with different permutations of the data, and reported the results by taking average of 10 runs. Please refer to the appendix for more details.

\begin{table}[!htbp]

\scriptsize
\centering
% \caption{Memory buffer capacity used for various datasets.}
% \label{Tabel_buffer_capacity}
\begin{tabular}{ c | c | c | c | c } %\hline
  \toprule
%   \multirow{2}{*} {\textbf{Parameters}} & \multicolumn{4}{c}{\textbf{Datasets}}  \\ \cmidrule{2-6}
    %  \\
    
  \textbf{Dataset} & \shortstack{CIFAR10} & \shortstack{CIFAR100} & \shortstack{ImageNet100} & \shortstack{iCubWorld 1.0}\\
    
  \midrule
    % \hline
  
    \shortstack{\textbf{Buffer} \\ \textbf{Capacity}} & 1000 & 1000 & 1000 & 180 \\
    
    \midrule
   
    \shortstack{\textbf{Training-Set} \\ \textbf{Size}} & 50000 & 50000 & 127778 & 6002 \\
   
 \bottomrule
\end{tabular}

\caption{Memory buffer capacity used for various datasets.}
\label{Tabel_buffer_capacity}

\end{table}

\subsection{Results}\label{results}

The detailed results of {\name} over various experimental settings along with the strong baseline methods are shown in Table~\ref{Table_results_main_paper}. We can clearly observe that {\name} consistently outperforms the baseline by a significant margin. The proposed model is also robust to the different streaming learning scenarios compared to the baselines. We repeat our experiment ten times, because of the space constraint, we omit standard-deviations and included in the appendix. We do not consider GDumb as the best-performing method, even when it achieves higher accuracy for the class-iid since it finetunes the network with the stored samples and makes the learning algorithm a \emph{`two-step'} process, which is prohibited in \emph{streaming-learning}.  We observe that `batch-learning' methods severely suffer from catastrophic forgetting. Moreover, replay-based `online-learning' method such as AGEM also suffer from information loss badly. Furthermore, GDumb, even with finetuning before each inference, cannot achieve the best accuracy in all the experiments.
CIFAR10/100, ImageNet are the standard classification datasets, while iCubWorld 1.0 is a challenging dataset that evaluates the models in more realistic scenarios or data-orderings. Particularly, class-instance ordering and instance ordering require the learner to learn from temporally ordered video frames one at a time. From Table~\ref{Table_results_main_paper}, we observe that {\name} obtain up to $\mathbf{8.58}\%$ and $\mathbf{6.26}\%$ improvement over the state-of-the-art streaming learning approaches. In fact, in most of the scenarios, {\name} is very close to the upper bound performance, i.e., when the model is trained in a fully offline fashion.

For completeness, we train \emph{`Our-method'} in `batch' and `online' learning setting to determine its effectiveness and compatibility in these settings. In Figure~\ref{fig:batch_vs_online_vs_streaming}, we compare $\boldsymbol{\Omega}_{\text{all}}$ results of various baseline with our-method {\name}. We observe that {\name} outperforms the baselines by a significant margin on both class-i.i.d and class-instance ordering on iCubWorld 1.0 dataset. This implies that a method trained in the extreme setting can be thought of as a universal method for all \emph{lifelong learning settings} with the widest possible applicability. More details are given in the appendix.

\section{Ablation Study}\label{ablation_study}

We perform extensive ablation to show the importance of the different components. The various ablation experiments validate the significance of the proposed components.

% \vspace{1mm}

% \subsection{Significance Of Different Sampling Strategies}

\textbf{Significance Of Different Sampling Strategies.} In Table~\ref{Table_different_sampling_ablation}, we compare the performance of {\name} while using various sampling strategies and memory replacement policies. We observe that for the buffer replacement, LAWRRR performs better compared to LAWCBR. Furthermore, for the sample replay, UAPN, along with LAWRRR memory buffer policy, outperforms other sampling strategies, except \emph{uniform sampling} (Uni) performs better on i.i.d ordering. We provide more details in the appendix.

\begin{table}[!htbp]
% \small
\scriptsize
\centering
% \caption{$\boldsymbol{\Omega}_{\text{all}}$ Results. For each experiment, the method with best performance is highlighted in \textbf{Bold}.}
% \label{Table_different_sampling_ablation}
\begin{tabular}{ c | c | c | c | c | c } %\hline
  \toprule

   \multirow{3}{*} {\shortstack{\textbf{Memory} \\ \textbf{Replacement}} } & \multirow{3}{*} {\shortstack{\textbf{Sample} \\ \textbf{Selection}}} & \multicolumn{2}{c}{\textbf{iCubWrold}} & \multicolumn{2}{c}{\textbf{ImageNet100}} \\
  
  \cmidrule{3-6}
    %  \\
 &  & \shortstack{\textbf{instance}} & \shortstack{\textbf{Class} \\ \textbf{instance}} & \shortstack{\textbf{iid}} & \shortstack{\textbf{Class-iid}}  \\
  \midrule

   \multirow{3}{*} { LAWCBR }  & Uni & 0.8975 & 0.8506 & 0.9582 & 0.9014 \\
   
  & UAPN & 0.9346 & 0.8500 &  0.9327  & 0.9135   \\
   
  & LAPN & 0.9172 & 0.8536 &  0.9253  & 0.9122   \\

  \cmidrule{1-6}

        \multirow{3}{*} { LAWRRR }  & Uni & 0.9269 & 0.9346 &  \textbf{0.9640}  & 0.8643  \\
   
  & UAPN & \textbf{0.9580} & \textbf{0.9585} &  0.9578 & \textbf{0.9171}  \\
   
  & LAPN & 0.9558 & 0.9497 & 0.9575 & 0.9112  \\

%     \midrule

%   & Offline & 1.0000 & 1.0000 & 1.0000 & 1.0000   \\
   
% %   \midrule
   
%   & $\widehat{\text{Offline}}$ & 0.7646 & 0.8840 & 0.8520 & 0.8953  \\
   
 \bottomrule
    % \hline
\end{tabular}

\caption{$\boldsymbol{\Omega}_{\text{all}}$ Results. For each experiment, the method with best performance is highlighted in \textbf{Bold}.}
\label{Table_different_sampling_ablation}

\end{table}

% \subsection{Choice of Hyperparameter ( $\boldsymbol{\lambda}_{2}$)}

\begin{figure}[!htp]
    \centering
    \includegraphics[width=8cm, height=2.7cm]{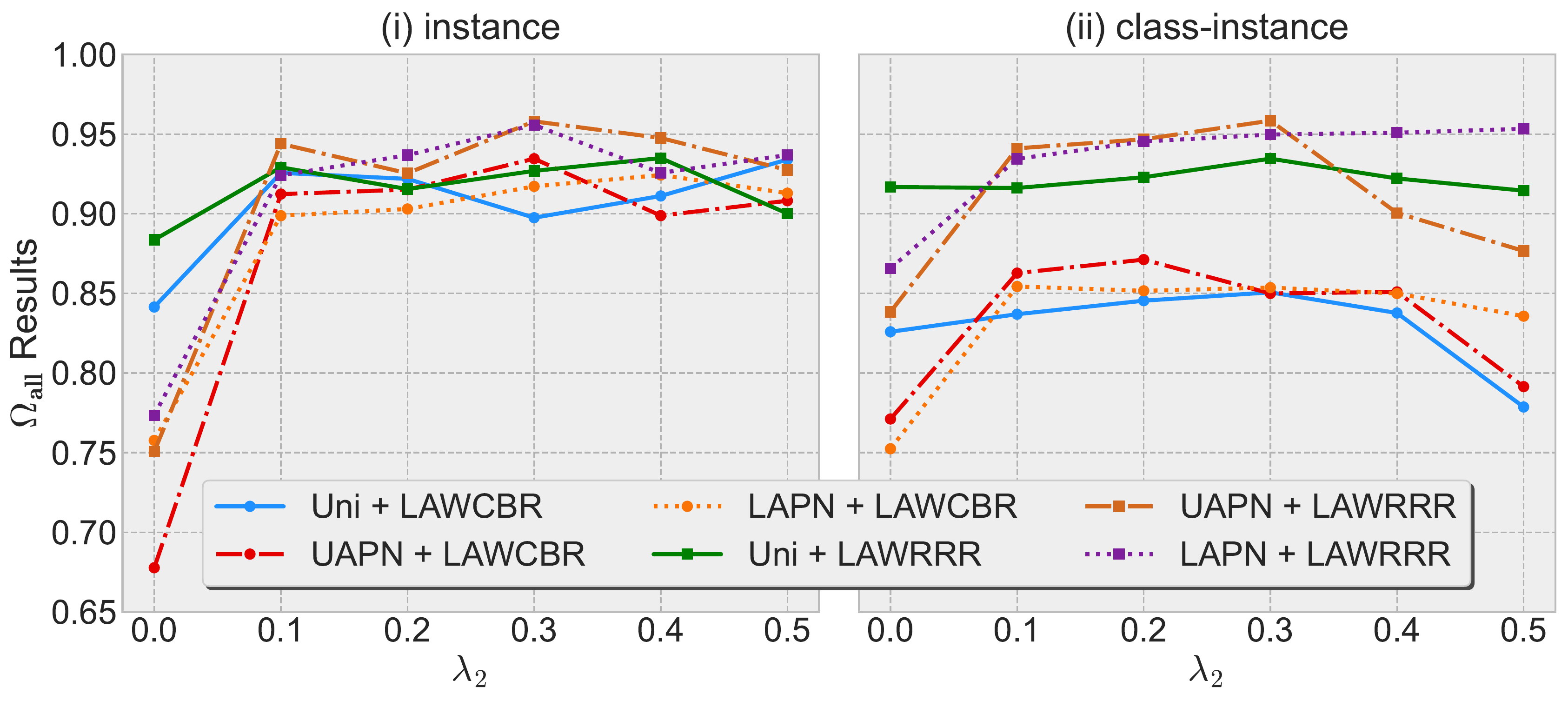}
    % \vspace{-5mm}
    \caption{Plots of $\boldsymbol{\Omega}_{\text{all}}$ as a function of hyper-parameter $\boldsymbol{\lambda}_{2}$ and different sampling strategies and replacement policies for (i) instance, (ii) class-instance ordering on iCubWorld 1.0.} 
    \label{fig:lambda_2_ablation_iCubWorld}
\end{figure}

\textbf{Choice Of Hyperparameter ( $\boldsymbol{\lambda}_{2}$).} Figure~\ref{fig:lambda_2_ablation_iCubWorld} shows the effect of changing the knowledge-distillation loss weight $\boldsymbol{\lambda}_{2}$ on the final $\boldsymbol{\Omega}_{\text{all}}$ accuracy for iCubWorld 1.0 on instance and class-instance ordering, while using different sampling strategies and buffer replacement policies. We observe the best model performance for $\boldsymbol{\lambda}_{2} = 0.3$, and we use this value of $\boldsymbol{\lambda}_{2}$ for all our experiments. We provide detailed ablation on $\boldsymbol{\lambda}_{2}$ in the appendix.

% \subsection{Significance of Knowledge-Distillation Loss}

\textbf{Significance Of Knowledge-Distillation Loss.}  Figure~\ref{fig:lambda_2_ablation_iCubWorld} with $\boldsymbol{\lambda}_{2} = 0.0$ represents the model without knowledge distillation. We can observe that the model performance significantly degrades without knowledge distillation. Therefore, knowledge distillation is a key component to the model's performance. More details are given in the appendix.

% \subsection{Choice Of Buffer Capacity}

\begin{figure}[!htp]
    \centering
    \includegraphics[width=6cm, height=2.7cm]{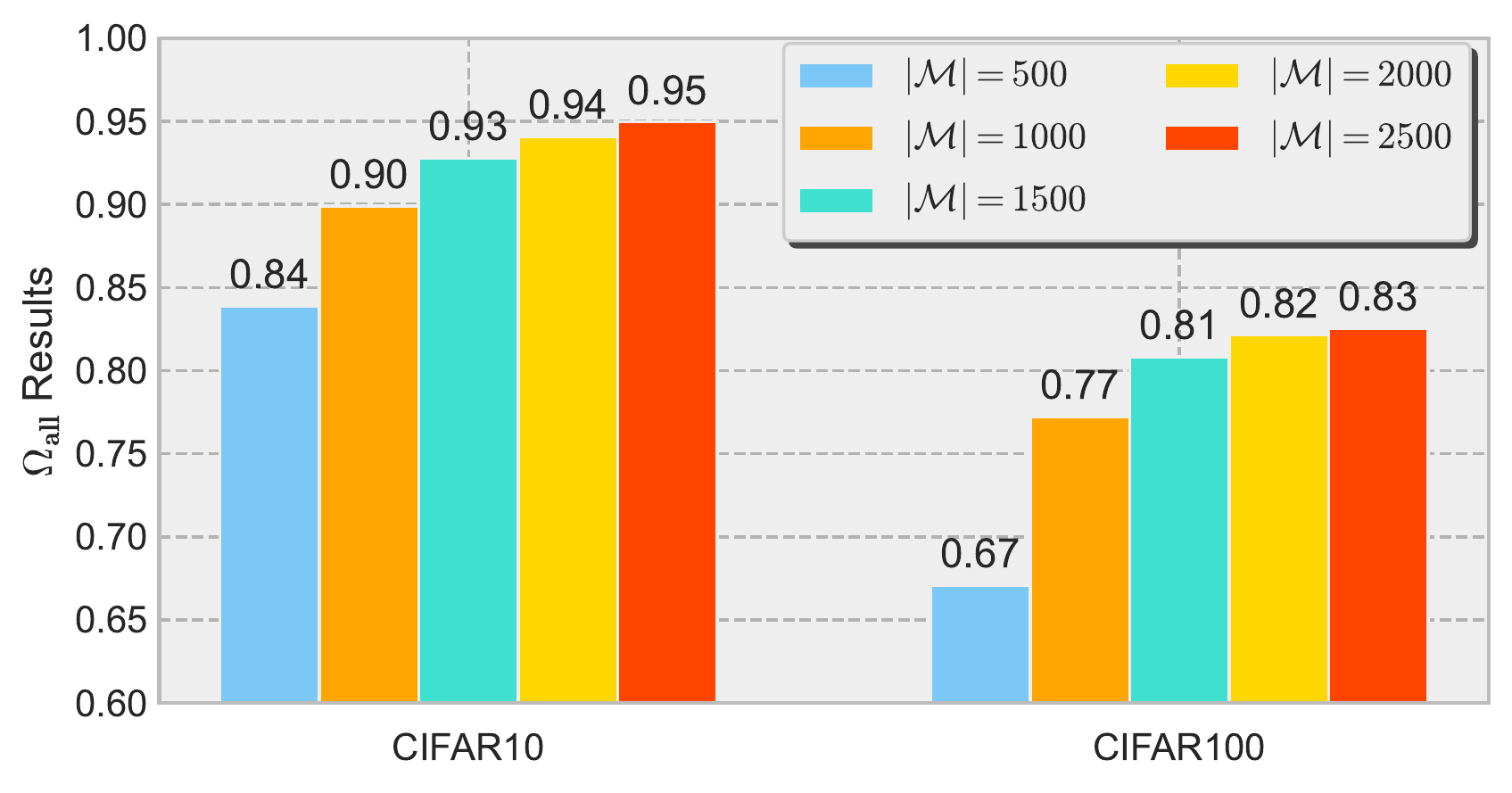}
    % \vspace{-5mm}
    \caption{Plots of $\boldsymbol{\Omega}_{\text{all}}$ as a function of buffer capacity $|\mathcal{M}|$ for class-i.i.d data-ordering on CIFAR10 and CIFAR100.} 
    \label{fig:memory_ablation_study}
    % \vspace{-4mm}
\end{figure}

\textbf{Choice Of Buffer Capacity.} We perform an ablation for the different buffer capacities, i.e., $|\mathcal{M}|$. The results are shown in Figure~\ref{fig:memory_ablation_study}. It is evident that, with the longer sequence of incoming data, the model's ({\name}) performance improves with the increase in the buffer capacity, as it helps minimize the confusion in the output prediction.

\section{Conclusion}
\emph{Streaming continual learning} (SCL) is the most challenging and realistic framework for continual learning; most of the recent promising models for the CL are unable to handle this above setting. Our work proposes a dual regularization and loss-aware sample replay to handle the SCL scenario. The proposed model is highly efficient since it learns a joint likelihood from the current and replay samples without leveraging any external finetuning. Also, to improve the training efficiency further, the proposed model selects a few most informative samples from the buffer instead of using the entire buffer for the replay. We have conducted a rigorous experiment over several challenging datasets and showed that {\name} outperforms state-of-the-art approaches in this setting by a significant margin. To disentangle the importance of the various components, we perform extensive ablation studies and observe that the proposed components are essential to handle the SCL setting.

%%%%%%%%% REFERENCES
{\small
\bibliographystyle{ieee_fullname}
\bibliography{egbib}
}

%%%%%%% Appendix

\clearpage

\appendix

\section{Preliminaries}\label{preliminaries}

\subsection{`Class Incremental Learning' V/S `Task Incremental/Multi-Task Learning'}\label{class_incremental_vs_multi_task_learning}

\emph{`Class incremental learning'}~\cite{rebuffi2017icarl, chaudhry2018riemannian, hayes2019memory, hayes2019remind}, is a challenging variant of lifelong learning, where the classifier needs to learn to discriminate between different class labels from different tasks. The key distinction between \emph{`class incremental learning'} and \emph{`task incremental/multi-task learning'}~\cite{kirkpatrick2017overcoming, aljundi2018memory,aljundi2018selfless, nguyen2017variational}, lies in how the classifier's accuracy is evaluated at the test time. In \emph{`class incremental learning'}, at the test time, the task identifier $t$ is not specified, and the accuracy is computed over all the observed classes with $\frac{1}{\mathcal{C}}$ chance, where $\mathcal{C}$ is the total number of classes accumulated so far. However, in \emph{`task incremental learning'}, the task identifier $t$ is known. 

For example, consider MNIST divided into $5$ tasks: $\left\{ \left\{ 0, 1 \right\}, \dots, \left\{ 8, 9 \right\}  \right\}$, which are used for sequential learning of a classifier. Then, at the end of $5$-th task, in \emph{`task incremental setting'}, the classifier needs to predict a class out of $\left \{ 8, 9 \right \}$ only. However, in \emph{`class incremental setting'}, a class label is predicted over all the ten classes that is observed so far, i.e., $\left\{ 0, \dots, 9 \right\}$ with $\frac{1}{10}$ chance for each class.

% \subsection{Bayesian Neural Networks}

\subsection{Variational Continual Learning (VCL)}\label{vcl}

Variational Continual Learning (VCL)~\cite{nguyen2017variational} is a recently proposed continual learning approach that mitigates catastrophic forgetting in neural networks in a Bayesian framework. It sets the posterior of parameters distribution as the prior before training on the next task, i.e., $p_{t}(\boldsymbol{\theta}) = q_{t - 1}(\boldsymbol{\theta})$, the new task reuses the previous task's posterior as the new prior. VCL solves the following KL divergence minimization problem while training on task $t$ with the new data $\mathcal{D}_{t}$:
\begin{equation}\label{eq_1}
    q_{t}(\boldsymbol{\theta}) = \argmin_{q \epsilon \mathcal{Q}} \text{KL} \left ( q(\boldsymbol{\theta}) \; \left|\right| \; \frac{1}{Z_{t}} q_{t - 1}(\boldsymbol{\theta}) \; p(\mathcal{D}_{t} | \boldsymbol{\theta}) \right )
\end{equation}

While offering a principled way of continual learning, VCL follows \emph{task incremental / multi task learning} setting, and uses \emph{`task specific head networks'}, for each task $t$, such that, $p(\boldsymbol{\theta} | \mathcal{D}_{1:t}) = p(\boldsymbol{\theta}_{t} | \mathcal{D}_{t}) p(\boldsymbol{\theta}_{S} | \mathcal{D}_{1:t})$, where $\boldsymbol{\theta} = \left \{ \boldsymbol{\theta}_{S}, \boldsymbol{\theta}_{t} \right \}$, $\boldsymbol{\theta}_{s}$ is shared between all the tasks, whereas $\boldsymbol{\theta}_{t}$ kept fixed after training on task $t$. This configuration prohibits knowledge transfer across tasks, and results in a poor accuracy in \emph{class incremental setting}~\cite{farquhar2018towards} for both VCL with or without Coreset.

VCL with Coreset~\cite{nguyen2017variational} withheld some data points from the task data before training and keeps them in a coreset. These data points are not used for the network training and are only used for finetuning the network before each inference. However, in \emph{online streaming learning} finetuning the network at any time is prohibited, as it makes the training process a \emph{two-step} learning process instead of \emph{single-pass} learning. Furthermore, the coreset is created by sampling data points from the entire task data, whereas in \emph{online streaming setting}, each instance arrives one at a time. Finally, the performance of \emph{VCL with Coreset} is heavily dependent on the finetuning with the withheld samples, i.e., coreset samples, before inference~\cite{farquhar2018towards}, and still not comparable enough to our proposed method ({\name}).

\subsection{REMIND}\label{remind}

REMIND~\cite{hayes2019remind} is a recently proposed replay-based lifelong learning approach which combats catastrophic forgetting~\cite{french1999catastrophic} in deep neural network in \emph{online-streaming setting}. While following such a challenging setting, it separates the convolutional neural network into two networks: $(i)$ a frozen feature extractor and $(ii)$ a plastic neural network. Learning involves the following steps: $(i)$ compression of each new input using product quantization (PQ)~\cite{jegou2010product}, $(ii)$ reconstruction of the previously stored compressed representations using PQ, and $(iii)$ mixing the reconstructed past examples with the new input and updating the parameters of the plastic layers of the network.

While it offers a principled way to combat catastrophic forgetting and achieves state-of-the-art performance, there are few concerns that can be limiting in the continual learning setup. It stores considerably a large number of past examples compared to the baselines; for example, iCaRL~\cite{rebuffi2017icarl} stores 10K past examples for ImageNet experiment whereas REMIND stores 1M past examples. Furthermore, REMIND actually uses a lossy compression method (PQ) to store the past samples, which is merely an engineering technique far from any algorithmic improvement and can be used by any lifelong learning approach.

\subsection{Bayesian Neural Network}

Bayesian neural networks~\cite{neal2012bayesian} are discriminative models, which extend the standard deep neural networks with Bayesian inference. The network parameters are assumed to have a prior distribution, $p(\boldsymbol{\theta})$, and it infers the posterior given the observed data $\mathcal{D}$, that is, $p(\boldsymbol{\theta} | \mathcal{D})$. However, the exact posterior inference is computationally intractable for any complex models, and an approximation is needed. One such scheme is \emph{`Bayes-by-Backprop'}~\cite{blundell2015weight}. It uses a mean-field variational posterior $q(\boldsymbol{\theta})$ over the network parameters and uses reparameterization-trick~\cite{kingma2013auto} to sample from the posterior, which are then used to approximate the evidence lower bound (ELBO) via Monte-Carlo sampling.

% Our goal in \emph{`online streaming learning'} is to learn a Bayesian neural network, which can learn from a sequence of data coming one at time by inferring the posterior $q_{t}(\boldsymbol{\theta})$ for each time instance $t$, without forgetting the information contained in the posterior inferred from the past data.

In our proposed method ({\name}), we have used a Bayesian neural network (BNN) as the plastic network $F(\cdot)$. We have discussed training the plastic network (BNN) $F(\cdot)$ with a single step posterior update without catastrophic forgetting~\cite{french1999catastrophic} in Section 3.1 in the main paper.

%%%%%%%%%%%%%%%%%%%%%%%%%%%%%%%%%%%%%%%%%%%%%%%%%%%%%%%%%%%%
%% Compare batch vs online vs streaming, main paper results with standard deviations
%%%%%%%%%%%%%%%%%%%%%%%%%%%%%%%%%%%%%%%%%%%%%%%%%%%%%%%%%%%%

\begin{table*}[!htbp]
  % \small
  \scriptsize
  \centering

   \resizebox{\textwidth}{!}{
  
  \begin{tabular}{ c | c | c | c | c | c | c } %\hline
    \toprule

    \multirow{3}{*} { \textbf{Method} } & \multicolumn{6}{c}{\textbf{iCubWorld 1.0}}  \\
    
    \cmidrule{2-7}
  
      & \multicolumn{3}{c}{\textbf{class-iid}} & \multicolumn{3}{c}{\textbf{class-instance}} \\
      
      \cmidrule{2-7}
      
      & \textbf{Batch} & \textbf{Online} & \textbf{Streaming} & \textbf{Batch} & \textbf{Online} & \textbf{Streaming} \\
      
      \midrule
      
      VCL & 0.5493 $\pm$ 0.0372 & 0.4835 $\pm$ 0.0132 & 0.3806 $\pm$ 0.0527 & 0.3299 $\pm$ 0.0469 & 0.3469 $\pm$ 0.0031 & 0.3473 $\pm$ 0.0025 \\
      
      \shortstack{VCL w/ C/} & 0.5314 $\pm$ 0.0306 & 0.4849 $\pm$ 0.0093 & 0.3948 $\pm$ 0.0558 & 0.4353 $\pm$ 0.0354 & 0.4373 $\pm$ 0.0284 & 0.4705 $\pm$ 0.0165 \\

      A-GEM & - & 0.4890 $\pm$ 0.0063 & 0.4047 $\pm$ 0.0632 & - & 0.3497 $\pm$ 0.0013 & 0.3489 $\pm$ 0.0030 \\
      
      %GDumb & - & \textbf{0.9660 $\pm$ 0.0201} & - & - & 0.7908 $\pm$ 0.0329 & - \\
      
      TinyER & 0.9109 $\pm$ 0.0241 & 0.8382 $\pm$ 0.0332 & 0.9069 $\pm$ 0.0297 & 0.8106 $\pm$ 0.0486 & 0.7042 $\pm$ 0.0526 & 0.8215 $\pm$ 0.0341 \\
      
      REMIND & 0.8381 $\pm$ 0.0333 & 0.6525 $\pm$ 0.0426 & 0.8553 $\pm$ 0.0349 & 0.6170 $\pm$ 0.0930 & 0.5879 $\pm$ 0.0473 & 0.7615 $\pm$ 0.0319 \\
      
      % \midrule
      
      \textbf{Ours} & \textbf{0.9585 $\pm$ 0.0184} & \textbf{0.8969 $\pm$ 0.0320} & \textbf{0.9480 $\pm$ 0.0215} & \textbf{0.8769 $\pm$ 0.0647} & \textbf{0.8300 $\pm$ 0.0587} & \textbf{0.9585 $\pm$ 0.0223} \\
      
      % \midrule
      
      % Offline & 1.0000 & 1.0000 & 1.0000 & 1.0000 & 1.0000 & 1.0000 \\
      
      % % \midrule
      
      % $\widehat{\text{Offline}}$ & 0.8849 & 0.8849 & 0.8849  & 0.8840 & 0.8840 & 0.8840 \\

   \bottomrule
      % \hline
  \end{tabular}

   }
  
  \caption{$\boldsymbol{\Omega}_{\text{all}}$ with their associated standard deviations for \emph{`batch'}, \emph{`online'}, and \emph{`streaming'} versions of baselines on iCubWorld 1.0 on $(i)$ class-i.i.d, and $(ii)$ class-instance ordering. `-' indicates experiments we were unable to run, because of compatibility issues.}
  \label{Table_compare_incremental_vs_online_vs_streaming}
  
\end{table*}

%%%%%%%%%%%%%%%%%%%%%%%%%%%%%%%%%%%%%%%%%%%%%%%%%%%
%% main paper results with standard deviations
%%%%%%%%%%%%%%%%%%%%%%%%%%%%%%%%%%%%%%%%%%%%%%%%%%%

\begin{table*}[!htbp]
  \scriptsize
  \centering

   \resizebox{\textwidth}{!}{
%   \scriptsize
  \begin{tabular}{ c | c c c | c c c } %\hline
    \toprule
    \multirow{2}{*} {\textbf{Method}} & \multicolumn{3}{c}{\textbf{iid}} & \multicolumn{3}{c}{\textbf{Class-iid}} \\ 
    
    \cmidrule{2-7}

    & \shortstack{\textbf{CIFAR10}} & \shortstack{\textbf{CIFAR100}} & \shortstack{\textbf{ImageNet100}} & \shortstack{\textbf{CIFAR10}} & \shortstack{\textbf{CIFAR100}} & \shortstack{\textbf{ImageNet100}} \\

    \midrule 
      
      Fine-Tune & 0.1175 $\pm$ 0.0000 & 0.0180 $\pm$ 0.0035 & 0.0127 $\pm$ 0.0029 & 0.3447 $\pm$ 0.0003  & 0.1277 $\pm$ 0.0022 & 0.1223 $\pm$ 0.0052  \\
      
      EWC & - & - & - & 0.3446 $\pm$ 0.0003 & 0.1292 $\pm$ 0.0037 & 0.1225 $\pm$ 0.0039 \\
      
      MAS & - & - & - & 0.3470 $\pm$ 0.0075  & 0.1280 $\pm$ 0.0029 & 0.1234 $\pm$ 0.0046 \\
      
      VCL & - & - & - & 0.3442 $\pm$ 0.0006  & 0.1273 $\pm$ 0.0041 & 0.1205 $\pm$ 0.0015  \\
      
      \textcolor{red}{VCL w/ C/} & - & - & - & 0.3716 $\pm$ 0.0501  & 0.1414 $\pm$ 0.0224 & 0.1259 $\pm$ 0.0122 \\
      
      \textcolor{red}{Coreset}  & - & - & - & 0.3684 $\pm$ 0.0442  & 0.1432 $\pm$ 0.0256 & 0.1273 $\pm$ 0.0182  \\
      
      \textcolor{red}{GDumb} & 0.8686 $\pm$ 0.0065 & 0.6067 $\pm$ 0.0119 & 0.8361 $\pm$ 0.0070 & \textit{\textbf{0.9252 $\pm$ 0.0057}}  & 0.7635 $\pm$ 0.0096 & \textit{\textbf{0.9197 $\pm$ 0.0081}}  \\
      
      A-GEM & 0.1175 $\pm$ 0.0000 & 0.0182 $\pm$ 0.0035 & 0.0139 $\pm$ 0.0041 & 0.3448 $\pm$ 0.0002  & 0.1290 $\pm$ 0.0037 & 0.1215 $\pm$ 0.0025 \\
      
      TinyER & 0.9314 $\pm$ 0.0114 & 0.7588 $\pm$ 0.0128 & 0.9415 $\pm$ 0.0085 & 0.8926 $\pm$ 0.0158  & 0.7402 $\pm$ 0.0195 & 0.8995 $\pm$ 0.0122  \\
      
      ExStream & 0.8866 $\pm$ 0.0244 & 0.7845 $\pm$ 0.0121 & 0.9293 $\pm$ 0.0082 & 0.8123 $\pm$ 0.0209  & 0.7176 $\pm$ 0.0208 & 0.8757 $\pm$ 0.0148 \\
      
      % Deep-SLDA &  &  &  &   &  &  &  &  &  &   \\
      
      REMIND & 0.8910 $\pm$ 0.0073 & 0.6457 $\pm$ 0.0091 & 0.9088 $\pm$ 0.0109 & 0.8832 $\pm$ 0.0201  & 0.6787 $\pm$ 0.0215 & 0.8803 $\pm$ 0.0157 \\
      
      % \midrule
      
      \textbf{Ours} & \textbf{0.9579 $\pm$ 0.0040} & \textbf{0.8679 $\pm$ 0.0057} & \textbf{0.9640 $\pm$ 0.0060} & \textbf{0.8991 $\pm$ 0.0089}  & \textbf{0.7724 $\pm$ 0.0188} & \textbf{0.9171 $\pm$ 0.0073}  \\
     
    \midrule
     
    Offline & 1.0000 & 1.0000 & 1.0000 & 1.0000  & 1.0000 & 1.0000 \\
     
  %   \midrule
     
    $\widehat{\text{Offline}}$ & 0.8509 & 0.6083 & 0.8520 & 0.8972  & 0.7154 & 0.8953 \\ 
     
   \bottomrule
   
      % \hline
  \end{tabular}
  }
  
  \caption{$\boldsymbol{\Omega}_{\text{all}}$ results with thier associated standard deviations. For each experiment, the method with best performance in \emph{`streaming-setting'} is highlighted in \textbf{Bold}. The reported results are average over $10$ runs with different permutations of the data. Offline model is trained only once. ${\widehat{\text{Offline}} =  \frac{1}{T}{\sum_{t = 1}^{T}{\boldsymbol{\alpha}_{\text{offline}, t}}}}$, where $T$ is the total number of testing events. `-' indicates
  experiments we were unable to run, because of compatibility issues. Methods in \textcolor{red}{Red} use fine-tuning before inference, which violates \emph{`single-pass'} learning constraint.}
  \label{Table_results_std_main_paper}
  
  \end{table*}

  \begin{table*}[!htbp]
  \centering
  \small
  
  \addtolength{\tabcolsep}{16.0pt}
  \begin{tabular}{ c | c | c | c | c } %\hline
    \toprule
    
    \multirow{2}{*} {\textbf{Method}} & \multicolumn{4}{c}{\textbf{iCubWorld 1.0}} \\ 
    
    \cmidrule{2-5}

    & \shortstack{\textbf{iid}}  & \shortstack{\textbf{Class-iid}} & \shortstack{\textbf{Instance}} & \shortstack{\textbf{Class-instance}} \\

    \midrule 
      
      Fine-Tune & 0.1369 $\pm$ 0.0184 & 0.3893 $\pm$ 0.0534  & 0.1307 $\pm$ 0.0000 & 0.3485 $\pm$ 0.0022 \\
      
      EWC & - & 0.3790 $\pm$ 0.0419 & - & 0.3487 $\pm$ 0.0034 \\
      
      MAS & - & 0.3912 $\pm$ 0.0613  & - & 0.3486 $\pm$ 0.0019 \\
      
      VCL & - & 0.3806 $\pm$ 0.0527  & - & 0.3473 $\pm$ 0.0025 \\
      
      \textcolor{red}{VCL w/ C/} & - & 0.3948 $\pm$ 0.0558 & - & 0.4705 $\pm$ 0.0165 \\
      
      \textcolor{red}{Coreset}  & - & 0.3994 $\pm$ 0.0922  & - & 0.4669 $\pm$ 0.0251 \\
      
      \textcolor{red}{GDumb} & 0.8993 $\pm$ 0.0413 & \textit{\textbf{0.9660 $\pm$ 0.0201}}  & 0.6715 $\pm$ 0.0540 & 0.7908 $\pm$ 0.0329 \\
      
      A-GEM & 0.1311 $\pm$ 0.0000 & 0.4047 $\pm$ 0.0632 & 0.1309 $\pm$ 0.0003 & 0.3489 $\pm$ 0.0030 \\
      
      TinyER & 0.9590 $\pm$ 0.0378 & 0.9069 $\pm$ 0.0297 & 0.8726 $\pm$ 0.0649 & 0.8215 $\pm$ 0.0341 \\
      
      ExStream & 0.9235 $\pm$ 0.0584 & 0.8820 $\pm$ 0.0285  & 0.8954 $\pm$ 0.0542 & 0.8727 $\pm$ 0.0229 \\
      
      % Deep-SLDA &  &  &  &   &  &  &  &  &  &   \\
      
      REMIND & 0.9260 $\pm$ 0.0311 & 0.8553 $\pm$ 0.0349  & 0.8157 $\pm$ 0.0600 & 0.7615 $\pm$ 0.0319 \\
      
      % \midrule
      
      \textbf{Ours} & \textbf{0.9716 $\pm$ 0.0141} & \textbf{0.9480 $\pm$ 0.0215}  & \textbf{0.9580 $\pm$ 0.0298} & \textbf{0.9585 $\pm$ 0.0223} \\
     
    \midrule
     
    Offline & 1.0000 & 1.0000 & 1.0000 & 1.0000  \\
     
  %   \midrule
     
    $\widehat{\text{Offline}}$ & 0.7626 & 0.8849  & 0.7646 & 0.8840 \\ 
     
   \bottomrule
   
      % \hline
  \end{tabular}

  \caption{$\boldsymbol{\Omega}_{\text{all}}$ results with thier associated standard deviations. For each experiment, the method with best performance in \emph{`streaming-setting'} is highlighted in \textbf{Bold}. The reported results are average over $10$ runs with different permutations of the data. Offline model is trained only once. ${\widehat{\text{Offline}} =  \frac{1}{T}{\sum_{t = 1}^{T}{\boldsymbol{\alpha}_{\text{offline}, t}}}}$, where $T$ is the total number of testing events. `-' indicates
  experiments we were unable to run, because of compatibility issues. Methods in \textcolor{red}{Red} use fine-tuning before inference, which violates \emph{`single-pass'} learning constraint.}
  \label{Table_results_std_main_paper_icubworld}

\end{table*}

%%%%%%%%%%%%%%%%%%%%%%%%%%%%%%%%%%%%%%%%%%%%%%%%%%%%%%%%%%%

\section{{$\boldsymbol{\Omega}_{\text{all}}$} Results With Their Associated Standard Deviations}\label{main_paper_results_with_standard_deviations}

We repeated each experiment 10 times with different permutations of the data and reported the results by taking the average over 10 runs. However, due to space constraint, we could not include the $\boldsymbol{\Omega}_{\text{all}}$ results with their associated standard deviations in the main paper, which we provide here.

In Table~\ref{Table_compare_incremental_vs_online_vs_streaming}, we provide the detailed results (corresponding Figure 1 of main paper) with their associated standard deviations comparing {\name} and other baselines in different learning settings. It empirically shows that {\name} which is designed considering the extreme and most restrictive \emph{online streaming setting} can be thought of as a universal lifelong learning method with the widest possible applicability.

Table~\ref{Table_results_std_main_paper} and Table~\ref{Table_results_std_main_paper_icubworld} provides the detailed results (main paper Table 2) of {\name} with their associated standard deviations over various experimental settings along with the state-of-the-art baselines. We observe that {\name} is the best performing method throughout all the experiments. Particularly, in the challenging scenarios such as class-instance and instance ordering where the model needs to learn from temporally ordered image sequence, the proposed approach ({\name}) achieves $\mathbf{8.58\%}$ and $\mathbf{6.26\%}$ improvement over the state-of-the-art baselines.

\section{Baselines And Compared Methods In Detail}\label{baselines_and_compared_methods_in_detail}

The proposed approach ({\name}) follows \emph{`online streaming setting'}, to the best of our knowledge, recent works ExStream~\cite{hayes2019memory}, and REMIND~\cite{hayes2019remind} are the only method that trains a \emph{deep neural network} following our setting. We compared our approach ({\name}) against these strong baselines. In addition, we have compared various `batch' and `online' learning methods, which we describe below.

For a fair comparison, we follow a similar network structure throughout all the methods. We separate a convolutional neural network (CNN) into two networks: $(i)$ \emph{non-plastic} feature extractor $G(\cdot)$, and $(ii)$ plastic neural network $F(\cdot)$. For a given input image $\boldsymbol{x}$, the predicted class label is computed as: $y = F(G(\boldsymbol{x}))$. Across all the methods, we use the same initialization step for the feature extractor $G(\cdot)$ (discussed in Section 3.5 in the main paper) and keep it frozen throughout the \emph{streaming learning}. For all the methods, only the plastic network $F(\cdot)$ is trained with one sample at a time in \emph{streaming manner}. For details on the structure of the plastic network $F(\cdot)$ across baselines along with {\name}, please refer to Section 5.3 in the main paper. 

In the below, we describe the baselines which we have evaluated along with our proposed method ({\name}) in \emph{online streaming setting}:

\begin{enumerate}

    \item \textbf{EWC}~\cite{kirkpatrick2017overcoming}\textbf{:} It is a regularization-based incremental learning method, which penalizes any changes to the network parameters by the important weight measure, the diagonal of the Fisher information matrix.
    
    \item \textbf{MAS}~\cite{aljundi2018memory}\textbf{:} It is another regularization-based lifelong learning method, where the importance weight of the network parameters are estimated by measuring the magnitude of the gradient of the learned function.
    
    \item \textbf{VCL}~\cite{nguyen2017variational}\textbf{:} It uses variational inference (VI) with a Bayesian neural network to mitigate catastrophic forgetting, where it uses the previously learned posterior as the prior while learning incrementally with the sequentially coming data. For more details, please refer to Section~\ref{vcl}.
    
    \item \textbf{VCL with Coreset}~\cite{nguyen2017variational}\textbf{:} This method is the same as pure VCL as mentioned above, except, at the end of training on each task, the network is finetuned with the coreset samples. We adapted the coreset selection in \emph{online streaming setting} and stored data points in coreset in an \emph{online manner}.

    \item \textbf{Coreset Only}~\cite{farquhar2018towards}\textbf{:} This method is exactly similar to \emph{VCL with Coreset}~\cite{nguyen2017variational}, except the prior which is used for variational inference is the initial prior each time, i.e., it is not updated with the previous posterior before training on a new task.

    \item \textbf{GDumb}~\cite{prabhu2020gdumb}\textbf{:} It is an \emph{online learning} method. It stores data points with a greedy sampler and retrains the network from scratch each time with stored samples before inference.
    
    \item \textbf{A-Gem}~\cite{chaudhry2018efficient}\textbf{:} It is another \emph{online learning} approach. It uses past task data stored in memory to build an optimization constraint to be satisfied by each new update. If the gradient violates the constraint, then it is projected such that the constraint is satisfied.

    \item \textbf{TinyER}~\cite{chaudhry2019continual}\textbf{:} It stores past task data points in a tiny episodic memory and replays them with the current training data to enable continual learning. 
    
    \item \textbf{ExStream}~\cite{hayes2019memory}\textbf{:} It is an \emph{online streaming learning} method, which uses memory replay to enable continual learning. It maintains buffers of prototypes to store the input vectors. Once the buffer is full, it combines the two nearest prototypes in the buffer and stores the new input vector.
    
    \item \textbf{REMIND}~\cite{hayes2019remind}\textbf{:} Similar to ExStream, it is another \emph{streaming learning} method, which enables lifelong learning with memory replay. For more details on REMIND, please refer to Section~\ref{remind}.
    
    \item \textbf{Fine-tuning:} It is a streaming learning baseline and serves as the lower bound on the network's performance. In this scenario, the network parameters are fine-tuned with one instance through the whole dataset for a single epoch.
    
    \item \textbf{Offline:} It serves as the upper bound on the network's performance, where the network is trained in the traditional way; the complete dataset is divided into multiple batches, and the network loops over them multiple times.
    
\end{enumerate}

\begin{table*}[t]
   \small
%   \scriptsize
  \centering
  
  \addtolength{\tabcolsep}{12.0pt}
  \begin{tabular}{ c  c  c  c  c  c } %\hline
    \toprule

    \textbf{Method} & \shortstack{\textbf{Learning} \\ \textbf{Type}} & \shortstack{\textbf{Fine-tunes}} & \shortstack{\textbf{Violates Constraints Of} \\ \textbf{Streaming Learning}} & \textbf{Regularize}  & \textbf{Memory} \\
    
    \midrule
    
      EWC & Batch & \xmark & \xmark & \cmark & \xmark \\
      
      MAS & Batch & \xmark & \xmark & \cmark & \xmark \\
      
      VCL & Batch & \xmark & \xmark & \cmark & \xmark \\
      
      VCL w/ C/ & Batch & \textcolor{red}{\cmark} & \textcolor{red}{\cmark} & \cmark & \cmark \\
      
      Coreset & Online & \textcolor{red}{\cmark} & \textcolor{red}{\cmark} & \xmark & \cmark \\
      
      GDumb & Online & \textcolor{red}{\cmark} & \textcolor{red}{\cmark} & \xmark & \cmark \\
    
      TinyER & Online & \xmark & \xmark & \xmark & \cmark \\
      
      A-GEM & Online & \xmark & \xmark & \cmark & \cmark \\
      
      ExStream & Streaming & \xmark & \xmark & \xmark & \cmark \\
      
      % Deep-SLDA & Streaming & No & No \\
      
      REMIND & Streaming & \xmark & \xmark & \xmark & \cmark \\
      
      % \midrule
      
      \textbf{Ours} & Streaming & \xmark & \xmark & \cmark & \cmark \\

   \bottomrule
      % \hline
  \end{tabular}
  
  \caption{Categorization of the baseline approaches depending on the underlying simplifying assumptions they impose.}
  \label{Table_baseline_categorization}
  
\end{table*}

\textbf{Note:} In \emph{online streaming setting}, finetuning the network with the stored samples is prohibited, as it violates the \emph{single-pass} learning constraint. `VCL with Coreset', `Coreset Only', and `GDumb' finetune the network parameters before inference; therefore, these methods have an extra advantage compared to true \emph{streaming learning} approaches, and they violate the \emph{single-pass} learning constraint. Therefore these methods cannot be considered as the best-performing methods even when they achieve better final accuracy as these methods are not true \emph{streaming learning} method.

Table~\ref{Table_baseline_categorization} categorizes the baselines according to the underlying assumptions that they impose. For baselines which finetunes the network before inference and violates the constraint of \emph{streaming learning}, such as \emph{single pass learning} constraint, have been marked in \textcolor{red}{red} in the corresponding column.

%% Imagenet100 ablation study

\begin{table}[ht]

  \scriptsize
  \centering

  \begin{tabular}{ c | c | c | c } %\hline
    \toprule

   \multirow{2}{*} {\shortstack{\textbf{Memory} \\ \textbf{Replacement}} } & \multirow{2}{*} {\shortstack{\textbf{Sample} \\ \textbf{Selection}}} & \multicolumn{2}{c}{\textbf{Imagenet100}} \\
    
    \cmidrule{3-4}
      %  \\
     & & \shortstack{\textbf{iid}} & \shortstack{\textbf{Class-iid}}  \\
    \midrule

    \multirow{3}{*} { LAWCBR }  & Uni & 0.9582 $\pm$ 0.0037 & 0.9014 $\pm$ 0.0073 \\
      
   & UAPN &  0.9327 $\pm$ 0.0052 & 0.9135 $\pm$ 0.0081  \\
      
   & LAPN &  0.9253 $\pm$ 0.0115 & 0.9122 $\pm$ 0.0091  \\

    \cmidrule{1-4}

    \multirow{3}{*} { LAWRRR }  & Uni & \textbf{0.9640 $\pm$ 0.0060}  & 0.8643 $\pm$ 0.0127  \\
      
   & UAPN &  0.9578 $\pm$ 0.0035 & \textbf{0.9171 $\pm$ 0.0073}  \\
      
  & LAPN & 0.9575 $\pm$ 0.0047 & 0.9112 $\pm$ 0.0075  \\

      % \midrule

    % & & Offline & 1.0000 & 1.0000   \\

    % & & $\widehat{\text{Offline}}$ & 0.8520 & 0.8953  \\
      
    \bottomrule
      
  \end{tabular}
  
  \caption{$\boldsymbol{\Omega}_{\text{all}}$ Results as a function of different memory replacement policies and sample selection strategies for $(i)$ i.i.d, and $(ii)$ class-i.i.d ordering on ImageNet100. }
  \label{Table_ablation_imagenet100}
  
  \end{table}

\section{Ablation Study Additional Results}\label{ablation_study_additional_results}

In this section, we provide additional results for ablation studies, which we could not provide in the main paper due to space constraints.

\begin{itemize}

  \item \textbf{ImageNet100.} In Table~\ref{Table_ablation_imagenet100}, we compare the final $\boldsymbol{\Omega}_{\text{all}}$ accuracy of {\name} for $(i)$ i.i.d, and $(ii)$ class-i.i.d ordering on Imagenet100 dataset while using different memory replacement policy and past sample selection strategies.

  \item \textbf{CIFAR10/100.} Table~\ref{Table_ablation_cifar10_cifar100} compares the final $\boldsymbol{\Omega}_{\text{all}}$ accuracy of the proposed model ({\name}) for $(i)$ i.i.d and $(ii)$ class-i.i.d ordering on CIFAR10 and CIFAR100 respectively while using different values for the knowledge-distillation hyper-parameter $\boldsymbol{\lambda}_{2}$, and different memory replacement policies and various sample selection strategies.

  \item \textbf{iCubWorld 1.0.} Table~\ref{Table_ablation_iCubWorld_lawcbr} and Table~\ref{Table_ablation_iCubWorld_lawrrr} compares the final $\boldsymbol{\Omega}_{\text{all}}$ accuracy of {\name} for $(i)$ i.i.d, $(ii)$ class-i.i.d, $(iii)$ instance, and $(iv)$ class-instance ordering on iCubWorld 1.0 dataset while using different knowledge-distillation hyper-parameter $\boldsymbol{\lambda}_{2}$ and different sampling strategies. For memory replacement policy, Table~\ref{Table_ablation_iCubWorld_lawcbr} uses \emph{`loss-aware weighted class balancing replacement (LAWCBR)'} strategy, whereas Table~\ref{Table_ablation_iCubWorld_lawrrr} uses \emph{`loss-aware weighted random replacement with a reservoir (LAWRRR)'} strategy.

\end{itemize}

%% CIFAR10 and CIFAR100 ablation study

\begin{table*}[!htbp]
   \small
%   \scriptsize
  \centering

  \addtolength{\tabcolsep}{5.0pt}
  \begin{tabular}{ c | c | c | c | c | c | c } %\hline
    \toprule

    \multirow{2}{*} { $\lambda_{2}$ } & \multirow{2}{*} {\shortstack{\textbf{Memory} \\ \textbf{Replacement}} } & \multirow{2}{*} {\shortstack{\textbf{Sample} \\ \textbf{Selection}}} & \multicolumn{2}{c}{\textbf{iid}} & \multicolumn{2}{c}{\textbf{class-iid}} \\
    
    \cmidrule{4-7}
      %  \\
     & & & \shortstack{\textbf{CIFAR10}} & \shortstack{\textbf{CIFAR100}} & \shortstack{\textbf{CIFAR10}} & \shortstack{\textbf{CIFAR100}} \\
    \midrule

    \multirow{6}{*} { $0.2$ }     &  \multirow{3}{*} { LAWCBR }  & Uni & 0.9542 $\pm$ 0.0053 & 0.8135 $\pm$ 0.0054 & 0.8942 $\pm$ 0.0062 & 0.7343 $\pm$ 0.0131  \\
     
   & & UAPN & 0.9084 $\pm$ 0.0121 & 0.4760 $\pm$ 0.0136 & 0.8957 $\pm$ 0.0125 &  0.6448 $\pm$ 0.0257 \\
     
   & & LAPN & 0.8462 $\pm$ 0.0414 & 0.3834 $\pm$ 0.0335 & 0.8797 $\pm$ 0.0149 & 0.5332 $\pm$ 0.0310  \\

    \cmidrule{2-7}

           &  \multirow{3}{*} { LAWRRR }  & Uni & \textbf{0.9584 $\pm$ 0.0035} & 0.8617 $\pm$ 0.0091 & 0.8792 $\pm$ 0.0104 &  0.7221 $\pm$ 0.0149 \\
     
   & & UAPN & 0.9567 $\pm$ 0.0031 & 0.8366 $\pm$ 0.0107 & 0.8978 $\pm$ 0.0107 & 0.7589 $\pm$ 0.0185 \\
     
   & & LAPN & 0.9530 $\pm$ 0.0037 & 0.8273 $\pm$ 0.0141 & 0.8986 $\pm$ 0.0127 &  0.7478 $\pm$ 0.0191 \\

      \midrule

      \multirow{6}{*} { $0.3$ }     &  \multirow{3}{*} { LAWCBR }  & Uni & 0.9529 $\pm$ 0.0062 & 0.8134 $\pm$ 0.0077 & 0.8970 $\pm$ 0.0088 &  0.7369 $\pm$ 0.0106 \\
     
   & & UAPN & 0.9145 $\pm$ 0.0071 & 0.5096 $\pm$ 0.0088 & 0.8944 $\pm$ 0.0093 & 0.6836 $\pm$ 0.0231  \\
     
   & & LAPN & 0.9046 $\pm$ 0.0152 & 0.4376 $\pm$ 0.0220 & 0.8798 $\pm$ 0.0230 &  0.6275 $\pm$ 0.0291 \\

    \cmidrule{2-7}

           &  \multirow{3}{*} { LAWRRR }  & Uni & 0.9579 $\pm$ 0.0040 & \textbf{0.8679 $\pm$ 0.0057} & 0.8838 $\pm$ 0.0088 & 0.7307 $\pm$ 0.0122  \\
     
   & & UAPN & 0.9567 $\pm$ 0.0031 & 0.8542 $\pm$ 0.0066 & 0.8991 $\pm$ 0.0089 & \textbf{0.7724 $\pm$ 0.0188}  \\
     
   & & LAPN & 0.9538 $\pm$ 0.0044 & 0.8453 $\pm$ 0.0120 & \textbf{0.9024 $\pm$ 0.0116} & 0.7573 $\pm$ 0.0193  \\

      % \midrule

  %  & & REMIND & $0.8910 \pm 0.0073$ & $0.6457 \pm 0.0091$ & $0.8832 \pm 0.0201$ & $0.6787 \pm 0.0215$  \\
     
  %   \midrule
     
  %  & & Offline & 1.0000 & 1.0000 & 1.0000 & 1.0000   \\
     
  %   \midrule
     
  %  & & $\widehat{\text{Offline}}$ & 0.8509 & 0.6083 & 0.8972 &  0.7154  \\
     
   \bottomrule
      
  \end{tabular}
  
  \caption{$\boldsymbol{\Omega}_{\text{all}}$ Results as a function of knowledge-distillation hyper-parameter $\lambda_{2}$ and different memory replacement policies and sample selection strategies for $(i)$ i.i.d ordering, and $(ii)$ class-i.i.d ordering on CIFAR10 and CIFAR100 datasets.}
  \label{Table_ablation_cifar10_cifar100}
  
\end{table*}

% iCubWorld 1.0 ablation study: LACBS

\begin{table*}[!htbp]
   \small
%   \scriptsize
  \centering

  \addtolength{\tabcolsep}{11.0pt}
  \begin{tabular}{ c | c | c | c | c | c } %\hline
    \toprule

    \multirow{2}{*} { $\lambda_{2}$ } & \multirow{2}{*} {\shortstack{\textbf{Sample} \\ \textbf{Selection}}} & \multicolumn{4}{c}{\textbf{iCubWorld 1.0}} \\
    
    \cmidrule{3-6}
      %  \\
     & & \shortstack{\textbf{iid}} & \shortstack{\textbf{Class-iid}} & \shortstack{\textbf{Instance}} & \shortstack{\textbf{Class-instance}} \\
    \midrule

    \multirow{3}{*} { $0.0$ }  & Uni & 0.9431 $\pm$ 0.0418 & 0.9105 $\pm$ 0.0333 & 0.8414 $\pm$ 0.0541 & 0.8259 $\pm$ 0.0316  \\
     
   & UAPN & 0.8775 $\pm$ 0.0753 & 0.8863 $\pm$ 0.0529 & 0.6777 $\pm$ 0.0764 & 0.7711 $\pm$ 0.0574  \\
     
   & LAPN & 0.8975 $\pm$ 0.0697 & 0.8675 $\pm$ 0.0498 & 0.7576 $\pm$ 0.0739 & 0.7524 $\pm$ 0.0655  \\

      \midrule

      % \hline
    \multirow{3}{*} { $0.1$ }  & Uni & \textbf{0.9885 $\pm$ 0.0245} & 0.9163 $\pm$ 0.0237  & 0.9257 $\pm$ 0.0299 & 0.8369 $\pm$ 0.0329  \\
     
   & UAPN & 0.9781 $\pm$ 0.0318 & 0.9167 $\pm$ 0.0263 & 0.9124 $\pm$ 0.0525 & 0.8627 $\pm$ 0.0285  \\
     
   & LAPN & 0.9779 $\pm$ 0.0206 & 0.9224 $\pm$ 0.0332 & 0.8988 $\pm$ 0.0544 & 0.8543 $\pm$ 0.0288  \\
     
  %  & & UA & $0.8727 \pm 0.0222$ & - & - & -  \\
     
  %  & & LA & $0.7529 \pm 0.1284$ & - & - & -  \\
   
   \midrule
   
   \multirow{3}{*} { $0.2$ }  & Uni & 0.9841 $\pm$ 0.0178 & 0.9154 $\pm$ 0.0217 & 0.9219 $\pm$ 0.0333 &  0.8454 $\pm$ 0.0283 \\
     
   & UAPN & 0.9868 $\pm$ 0.0181 & 0.9293 $\pm$ 0.0306 & 0.9152 $\pm$ 0.0229 &  \textbf{0.8712 $\pm$ 0.0266} \\
     
   & LAPN & 0.9645 $\pm$ 0.0189 & 0.9310 $\pm$ 0.0227 & 0.9030 $\pm$ 0.0503 & 0.8516 $\pm$ 0.0400  \\

      \midrule
      
      \multirow{3}{*} { $0.3$ }   & Uni & 0.9777 $\pm$ 0.0264 & 0.9257 $\pm$ 0.0288  & 0.8975 $\pm$ 0.0454 & 0.8506 $\pm$ 0.0310  \\
     
   & UAPN & 0.9868 $\pm$ 0.0125 & 0.9309 $\pm$ 0.0355 & \textbf{0.9346 $\pm$ 0.0395} &  0.8500 $\pm$ 0.0363 \\
     
   &  LAPN & 0.9745 $\pm$ 0.0174 & \textbf{0.9352 $\pm$ 0.0266} & 0.9172 $\pm$ 0.0373 &  0.8536 $\pm$ 0.0343 \\

      \midrule
      
      \multirow{3}{*} { $0.4$ } & Uni & 0.9782 $\pm$ 0.0200 & 0.9278 $\pm$ 0.0295 & 0.9112 $\pm$ 0.0327 & 0.8377 $\pm$ 0.0292  \\
     
   &  UAPN & 0.9815 $\pm$ 0.0178 & 0.9160 $\pm$ 0.0464 & 0.8988 $\pm$ 0.0419 & 0.8509 $\pm$ 0.0350  \\
     
   &  LAPN & 0.9718 $\pm$ 0.0271 & 0.9325 $\pm$ 0.0401 & 0.9243 $\pm$ 0.0512 & 0.8499 $\pm$ 0.0650  \\

      \midrule
      
      \multirow{3}{*} { $0.5$ }  & Uni & 0.9742 $\pm$ 0.0183 & 0.8858 $\pm$ 0.1505 & 0.9341 $\pm$ 0.0350 & 0.7787 $\pm$ 0.2008 \\
     
   &  UAPN & 0.9692 $\pm$ 0.0197 & 0.8587 $\pm$ 0.2278 & 0.9082 $\pm$ 0.0758 &  0.7914 $\pm$ 0.2033 \\
     
   &  LAPN & 0.9725 $\pm$ 0.0184 & 0.9006 $\pm$ 0.0635 & 0.9129 $\pm$ 0.0467 &  0.8357 $\pm$ 0.0334 \\

      % \midrule

  %  & & REMIND & $0.9260 \pm 0.0311$ & $0.8553 \pm 0.0349$ & $0.8157 \pm 0.0600$ & $0.7615 \pm 0.0319$  \\
     
  %   \midrule
     
  %  & & Offline & 1.0000 & 1.0000 & 1.0000 & 1.0000   \\
     
  %   \midrule
     
  %  & & $\widehat{\text{Offline}}$ & 0.7626 & 0.8849 & 0.7646 &  0.8840  \\
     
   \bottomrule
      % \hline
  \end{tabular}
  
  \caption{$\boldsymbol{\Omega}_{\text{all}}$ Results as a function of knowledge-distillation hyper-parameter $\lambda_{2}$, and \emph{`loss-aware weighted class balancing replacement'} (LAWCBR) and different sampling strategies for $(i)$ i.i.d, $(ii)$ class-i.i.d, $(iii)$ instance, and $(iv)$ class-instance ordering on iCubWorld 1.0 dataset. }
  \label{Table_ablation_iCubWorld_lawcbr}
  
\end{table*}

%% iCubWorld 1.0 ablation study LACBRS

\begin{table*}[!htbp]
   \small
%   \scriptsize
  \centering
  
  \addtolength{\tabcolsep}{11.0pt}
  
  \begin{tabular}{ c | c | c | c | c | c } %\hline
    \toprule

    \multirow{2}{*} { $\lambda_{2}$ } & \multirow{2}{*} {\shortstack{\textbf{Sample} \\ \textbf{Selection}}} & \multicolumn{4}{c}{\textbf{iCubWorld 1.0}} \\
    
    \cmidrule{3-6}
      %  \\
      & & \shortstack{\textbf{iid}} & \shortstack{\textbf{Class-iid}} & \shortstack{\textbf{Instance}} & \shortstack{\textbf{Class-instance}} \\
    \midrule

    \multirow{3}{*} { $0.0$ }  & Uni & 0.9298 $\pm$ 0.0329 & 0.9063 $\pm$ 0.0396 & 0.8837 $\pm$ 0.0544 & 0.9168 $\pm$ 0.0312  \\
      
    &  UAPN & 0.9184 $\pm$ 0.0379 & 0.8818 $\pm$ 0.0396 & 0.7507 $\pm$ 0.0732 &  0.8384 $\pm$ 0.0675 \\
      
    &  LAPN & 0.9285 $\pm$ 0.0357 & 0.8912 $\pm$ 0.0430 & 0.7735 $\pm$ 0.0458 & 0.8657 $\pm$ 0.0521  \\

      \midrule

      % \hline
    \multirow{3}{*} { $0.1$ }  & Uni & \textbf{0.9830 $\pm$ 0.0207} & 0.9240 $\pm$ 0.0276  & 0.9292 $\pm$ 0.0344 & 0.9162 $\pm$ 0.0255  \\
      
    & UAPN & 0.9644 $\pm$ 0.0260 & 0.9368 $\pm$ 0.0228 & 0.9439 $\pm$ 0.0362 & 0.9411 $\pm$ 0.0224  \\
      
    & LAPN & 0.9541 $\pm$ 0.0280 & 0.9402 $\pm$ 0.0368 & 0.9241 $\pm$ 0.0401 & 0.9345 $\pm$ 0.0235  \\

    \midrule
    
    \multirow{3}{*} { $0.2$ }  & Uni & 0.9600 $\pm$ 0.0312 & 0.9351 $\pm$ 0.0315 & 0.9155 $\pm$ 0.0299 & 0.9229 $\pm$ 0.0284  \\
      
    & UAPN & 0.9640 $\pm$ 0.0236 & 0.9415 $\pm$ 0.0307 & 0.9254 $\pm$ 0.0331 & 0.9468 $\pm$ 0.0273  \\
      
    & LAPN & 0.9684 $\pm$ 0.0160 & 0.9382 $\pm$ 0.0361 & 0.9368 $\pm$ 0.0376 &  0.9454 $\pm$ 0.0263 \\

      \midrule
      
      \multirow{3}{*} { $0.3$ } & Uni & 0.9716 $\pm$ 0.0141  &  0.9118 $\pm$ 0.0344 & 0.9269 $\pm$ 0.0383 & 0.9346 $\pm$ 0.0191   \\
      
    & UAPN & 0.9454 $\pm$ 0.0239 & 0.9480 $\pm$ 0.0215 & \textbf{0.9580 $\pm$ 0.0298} & \textbf{0.9585 $\pm$ 0.0223}  \\
      
    & LAPN & 0.9667 $\pm$ 0.0174 & \textbf{0.9538 $\pm$ 0.0303} & 0.9558 $\pm$ 0.0304 &  0.9497 $\pm$ 0.0239 \\

      \midrule
      
      \multirow{3}{*} { $0.4$ }  & Uni & 0.9611 $\pm$ 0.0153  & 0.9243 $\pm$ 0.0524 & 0.9350 $\pm$ 0.0319 & 0.9222 $\pm$ 0.0403  \\
      
    & UAPN & 0.9647 $\pm$ 0.0257 & 0.9387 $\pm$ 0.0315 & 0.9476 $\pm$ 0.0264 & 0.9005 $\pm$ 0.1257  \\
      
    & LAPN & 0.9615 $\pm$ 0.0194 & 0.9323 $\pm$ 0.0421 & 0.9257 $\pm$ 0.0212 &  0.9509 $\pm$ 0.0323 \\

      \midrule
      
      \multirow{3}{*} { $0.5$ } & Uni & 0.9615 $\pm$ 0.0301  & 0.9391 $\pm$ 0.0268 & 0.9001 $\pm$ 0.0555 & 0.9145 $\pm$ 0.0649 \\
      
    &  UAPN & 0.9526 $\pm$ 0.0179 & 0.9390 $\pm$ 0.0267 & 0.9275 $\pm$ 0.0322 & 0.8766 $\pm$ 0.2320  \\
      
    &  LAPN & 0.9495 $\pm$ 0.0215 & 0.9085 $\pm$ 0.1230 & 0.9369 $\pm$ 0.0187 &  0.9533 $\pm$ 0.0248 \\

  %     \midrule

  %  & & REMIND & $0.9260 \pm 0.0311$ & $0.8553 \pm 0.0349$ & $0.8157 \pm 0.0600$ & $0.7615 \pm 0.0319$  \\
      
      % \midrule
      
    % & & Offline & $1.0000$ & $1.0000$ & $1.0000$ & $1.0000$   \\
      
  %   \midrule
      
    % & & $\widehat{\text{Offline}}$ & $0.7626$ & $0.8849$ & $0.7646$ &  $0.8840$  \\
      
    \bottomrule
      % \hline
  \end{tabular}
  
  \caption{$\boldsymbol{\Omega}_{\text{all}}$ Results as a function of knowledge-distillation hyper-parameter $\lambda_{2}$, and \emph{`loss-aware weighted random replacement with a reservoir'} (LAWRRR) and different sampling strategies for $(i)$ i.i.d, $(ii)$ class-i.i.d, $(iii)$ instance, and $(iv)$ class-instance ordering on iCubWorld 1.0 dataset. }
  \label{Table_ablation_iCubWorld_lawrrr}
  
\end{table*}

\begin{table*}[!htbp]
%   \scriptsize
  \small
  \centering
  
  \addtolength{\tabcolsep}{18.0pt}
  \begin{tabular}{ c | c | c | c | c} 
    \toprule
    \multirow{2}{*} {\textbf{Parameters}} & \multicolumn{4}{c}{\textbf{Datasets}}  \\ \cmidrule{2-5}
      
    & \shortstack{\textbf{CIFAR10}} & \shortstack{\textbf{CIFAR100}} & \shortstack{\textbf{ImageNet100}} & \shortstack{\textbf{iCubWorld 1.0}} \\
    \midrule 
      
    Optimizer & SGD & SGD & SGD & SGD \\
     
    Learning Rate & 0.01 & 0.01 & 0.01 & 0.01 \\
     
    Momentum & 0.9 & 0.9 & 0.9 & 0.9 \\
     
    Weight Decay & 1e-05 & 1e-05 & 1e-05 & 1e-05 \\
     
    Hidden Layer & [256, 256] & [256, 256] & [256, 256] & [256, 256]\\
     
    Activation & ReLU & ReLU & ReLU & ReLU\\
     
    Offline Batch Size & 128 & 128 & 256 & 16 \\
     
    Offline Epoch & 50 & 50 & 100 & 30 \\
     
    Buffer Capacity & 1000 & 1000 & 1000 & 180 \\
  
    \shortstack{Train-Set Size} & 50000 & 50000 & 127778 & 6002 \\
     
   \bottomrule
      
  \end{tabular}
  
  \caption{Training parameters used for {\name} and Offline model.}
  \label{Table_parameter_details}
  
\end{table*}

\section{ImageNet-100}\label{imagenet100}

In this paper, we used a subset of ImageNet-1000 (ILSVRC-2012)~\cite{russakovsky2015imagenet} that contains randomly chosen 100 classes. To ease a relevant study, we release the list of these 100 classes that we used to evaluate the streaming learner's performance in our experiments, as mentioned in Table~\ref{Table_imagenet100_list}.

\begin{table}[!htbp]
  \small
  % \scriptsize
  \centering

  \begin{tabular}{ c  c  c  c } %\hline
    \toprule
  
    \multicolumn{4}{c} {\textbf{List Of ImageNet-100 Classes}} \\

    \midrule

    n01632777 & n01667114 & n01744401 & n01753488 \\
    
    n01768244 & n01770081 & n01798484 & n01829413 \\

    n01843065 & n01871265 & n01872401 & n01981276 \\
    
    n02006656 & n02012849 & n02025239 & n02085620 \\

    n02086079 & n02089867 & n02091831 & n02094258 \\

    n02096294 & n02100236 & n02100877 & n02102040 \\

    n02105251 & n02106550 & n02110627 & n02120079 \\

    n02130308 & n02168699 & n02169497 & n02177972 \\

    n02264363 & n02417914 & n02422699 & n02437616 \\

    n02483708 & n02488291 & n02489166 & n02494079 \\

    n02504013 & n02667093 & n02687172 & n02788148 \\

    n02791124 & n02794156 & n02814860 & n02859443 \\

    n02895154 & n02910353 & n03000247 & n03208938 \\

    n03223299 & n03271574 & n03291819 & n03347037 \\

    n03445777 & n03529860 & n03530642 & n03602883 \\

    n03627232 & n03649909 & n03666591 & n03761084 \\

    n03770439 & n03773504 & n03788195 & n03825788 \\

    n03866082 & n03877845 & n03908618 & n03916031 \\

    n03929855 & n03954731 & n04009552 & n04019541 \\

    n04141327 & n04147183 & n04235860 & n04285008 \\

    n04286575 & n04328186 & n04347754 & n04355338 \\

    n04423845 & n04442312 & n04456115 & n04485082 \\

    n04486054 & n04505470 & n04525038 & n07248320 \\

    n07716906 & n07730033 & n07768694 & n07836838 \\

    n07860988 & n07871810 & n11939491 & n12267677 \\
     
   \bottomrule
      % \hline
  \end{tabular}
  
  \caption{The list of classes from ImageNet-100, which are randomly chosen from the original ImageNet-1000 (ILSVRC-2012)~\cite{russakovsky2015imagenet}.}
  \label{Table_imagenet100_list}
  
\end{table}

\section{Additional Implementation Details}\label{additional_implementation_details}

In this section, we provide some additional implementation details, which we could not provide in the main paper due to space constraints.

We use Mobilenet-V2~\cite{sandler2018mobilenetv2} pre-trained on ImageNet~\cite{russakovsky2015imagenet} available in PyTorch~\cite{paszke2019pytorch} TorchVision package as the base architecture for the feature extractor $G(\cdot)$. We use the convolutional base of Mobilenet-V2~\cite{sandler2018mobilenetv2} as the feature extractor $G(\cdot)$ to obtain embeddings from the raw pixels; we keep it frozen throughout the streaming learning. We use \emph{uniform sampling} and knowledge-distillation hyper-parameter $\lambda_{2} = 0.2$ for \emph{online learning} and \emph{batch learning} experiments (in Table~\ref{Table_compare_incremental_vs_online_vs_streaming}). We provide the parameter settings for the proposed method ({\name}) and the offline models in Table~\ref{Table_parameter_details}.

\section{Evaluation Over Different Data Orderings Additional Details}\label{evaluation_over_different_data_orderings}

The proposed approach ({\name}) is robust to various \emph{streaming learning scenarios} that can induce catastrophic forgetting~\cite{french1999catastrophic}. We evalaute the model's streaming learning ability with the four challenging data ordering~\cite{hayes2019remind, hayes2019memory} schemes: $(i)$ \emph{`streaming iid'}, $(ii)$ \emph{`streaming class iid'}, $(iii)$ \emph{`streaming instance'}, and $(iv)$ \emph{`streaming class instance'}. We described this four data ordering schemes in more detail in Section 5.2 in the main paper.

\textbf{Note:} Only iCubWorld 1.0~\cite{fanello2013icub} dataset contains the temporal ordering, therefore, \emph{`streaming instance'} \emph{`streaming class instance'} setting evaluated only on the iCubWorld 1.0 dataset.

In the below, we describe the following: $(i)$ how the base initialization is performed, and $(ii)$ how the network is trained in \emph{streaming setting} according to various data ordering schemes on different datasets.

% \begin{itemize}

\subsection{CIFAR10}

CIFAR10~\cite{krizhevsky2009learning} is a standard image classification dataset. It contains 10 classes with each consists of 5000 training images and 1000 testing images. Since, it does not contain any temporally ordered image sequence, we use CIFAR10 to evaluate the \emph{streaming learner's} ability in \emph{streaming i.i.d} and \emph{streaming class-i.i.d} orderings.

\begin{itemize}

  \item \textbf{\emph{streaming i.i.d:}} For the base initialization, we randomly select $2\%$ samples from the dataset and train the model in offline manner. Then we randomly shuffle the remaining samples and train the model incrementally with these samples by feeding one at a time in a streaming manner. 
  
  \item \textbf{\emph{streaming class-i.i.d:}} In base initialization, the model is trained in a typical offline mode with the samples from the first two classes. Then, in each incremental step, we select the samples from the next two classes, which are not included earlier. These samples are randomly shuffled and fed into the model in a streaming manner.

\end{itemize}

\subsection{CIFAR100}

CIFAR100~\cite{krizhevsky2009learning} is another standard image classification dataset. It contains 100 classes with each consists of 500 training images and 100 testing images. We use CIFAR100 to evaluate the model's ability in \emph{streaming i.i.d} and \emph{streaming class-i.i.d} orderings.

\begin{itemize}

  \item \textbf{\emph{streaming i.i.d:}} In this setting, we follow the similar approach as mentioned for the CIFAR10 dataset, with the only exception is that the base initialization is performed with $10\%$ randomly chosen samples, and the remaining samples are used for streaming learning.
  
  \item \textbf{\emph{streaming class-i.i.d:}} This approach also follows the similar approach as mentioned for the CIFAR10 dataset. However, in each incremental step, including the base initialization, we use samples from 10 classes. For the base initialization, we select samples from the first ten classes, and in each incremental step, we select samples from the succeeding ten classes which are not observed earlier.

\end{itemize}

\subsection{ImageNet100}

ImageNet100 is a subset of ImageNet-1000 (ILSVRC-2012)~\cite{russakovsky2015imagenet} that contains randomly chosen 100 classes, with each classes containing $700-1300$ training samples and $50$ validation samples. Since, for test samples, we do not have the ground truth labels, we use the validation data for testing the model's accuracy. We provide more details on ImageNet100 in Section~\ref{imagenet100}.

We use ImageNet100 dataset to evaluate the model's ability in \emph{streaming i.i.d} and \emph{streaming class-i.i.d} orderings.

\begin{itemize}

  \item \textbf{\emph{streaming i.i.d:}} In this case, we follow the similar approach as mentioned for CIFAR100 \emph{streaming i.i.d} ordering.
  
  \item \textbf{\emph{streaming class-i.i.d:}} We follow the similar approach as has been mentioned for CIFAR100 \emph{streaming class-i.i.d} ordering.

\end{itemize}

\begin{figure}[!htbp] 
	\centering
	\includegraphics[width=8.4cm, height=20.9cm]{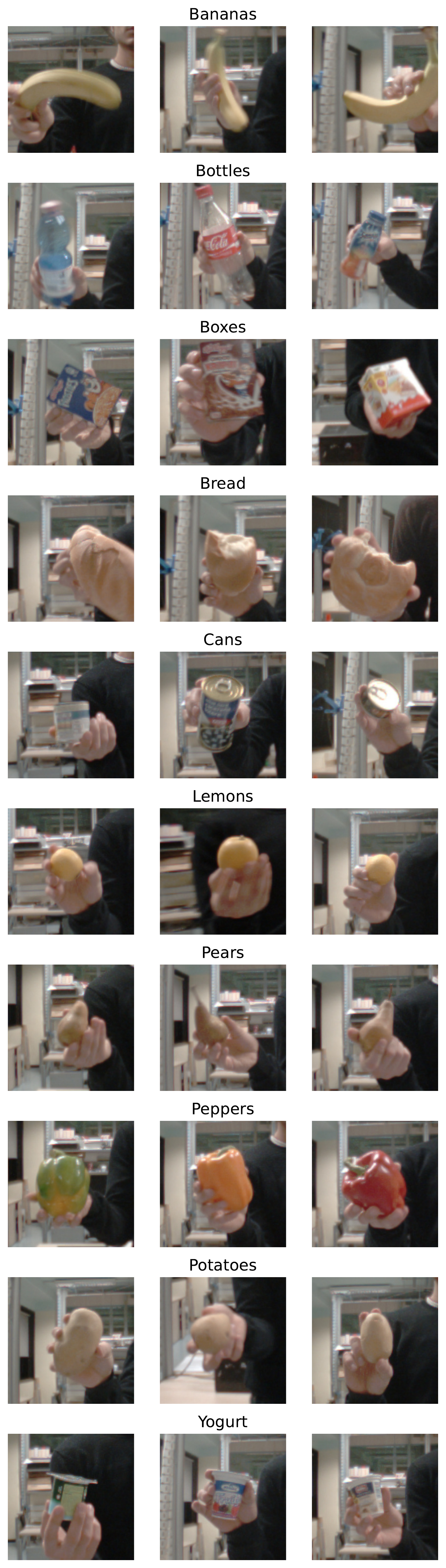}
	\caption{The iCubWorld $1.0$ \cite{fanello2013icub} dataset. $10$ categories: Bananas, Bottles, Boxes, Bread, Cans, Lemons, Pears, Peppers, Potatoes, Yogurt. Each category contains $3$ different instances.}
	\label{fig:iCubWorld_dataset_plot}
\end{figure}

\subsection{iCubWorld 1.0}

iCubWorld 1.0~\cite{fanello2013icub} is an object recognition dataset containing the sequence of video frames, with each frame containing only a single object. It is a more challenging and realistic dataset w.r.t the other standard datasets such as CIFAR10, CIFAR100, and ImageNet100. Technically, it is an ideal dataset to evaluate a model's performance in \emph{streaming learning} scenarios that are known to induce catastrophic forgetting~\cite{french1999catastrophic}, as it requires learning from temporally ordered image sequences, which are naturally \emph{non-i.i.d} images.

It contains 10 classes, each with 3 different object instances with $200-201$ images each. Overall, each class contains $600-602$ samples for training and $200-201$ samples for testing. Figure~\ref{fig:iCubWorld_dataset_plot} shows example images of the $30$ object instances in iCubWorld 1.0, where each row denotes one of the $10$ categories.

We use iCubWorld 1.0 to evaluate the performance of the streaming learner's in all the four data ordering schemes, i.e., $(i)$ \emph{streaming i.i.d}, $(ii)$ \emph{streaming class-i.i.d}, (iii) \emph{streaming instance}, and $(iv)$ \emph{streaming class-instance}.

\begin{itemize}

  \item \textbf{\emph{streaming i.i.d:}} In this setting, we follow the similar approach as mentioned for the CIFAR10 dataset, with the only exception, that is, $10\%$ randomly selected samples are used for the base initialization, and the rest are used for streaming learning.
  
  \item \textbf{\emph{streaming class-i.i.d:}} In this case, we follow the same strategy as mentioned for CIFAR10 \emph{streaming class-i.i.d} ordering.

  \item \textbf{\emph{streaming class-instance:}} In base initialization, the model is trained in a typical offline mode with the samples from the first two classes. In each incremental step, the network is trained in a streaming manner with the samples from the succeeding two classes which were not observed earlier. However, in this case, $(i)$ samples within a class are temporally ordered based on different object instances, and $(ii)$ all samples from one class are fed into the network before feeding any samples from the other class.
  
  \item \textbf{\emph{streaming instance:}} For the base initialization, $10\%$ randomly chosen samples are used, and the remaining samples are used to train the model incrementally with one sample at a time. In streaming setting, the samples are temporally ordered based on different object instances. Specifically, we organize the data stream by putting temporally ordered $50$ frames of an object instance, then we put temporally ordered $50$ frames of the second object instance, and so on. In this way, after putting $50$ temporally ordered frames from each object instance, we put the next $50$ temporally ordered frames of the first object instance and follow the earlier approach until all the frames of each instance have been exhausted.

\end{itemize}

\section{Derivation of Joint Posterior}\label{derivation_of_joint_posterior}

\small
\begin{align*}
\mathcal{L}^{1}_{t}(\boldsymbol{\theta}) & = \argmin_{q \epsilon \mathcal{Q}} \text{KL}\left[ q_{t}(\boldsymbol{\theta}) \left|\right| \frac{1}{Z_{t}} q_{t - 1}(\boldsymbol{\theta}) p(\mathcal{D}_{t} | \boldsymbol{\theta}) p(\mathcal{D}_{\mathcal{M}, t} | \boldsymbol{\theta})  \right]\\
& \backsimeq \argmin_{q \epsilon \mathcal{Q}} \text{KL}\left[ q_{t}(\boldsymbol{\theta}) || q_{t - 1}(\boldsymbol{\theta}) p(\mathcal{D}_{t} | \boldsymbol{\theta}) p(\mathcal{D}_{\mathcal{M}, t} | \boldsymbol{\theta})  \right]\\
&=\int q_{t}(\boldsymbol{\theta}) \log \frac{q_{t}(\boldsymbol{\theta})}{q_{t - 1}(\boldsymbol{\theta}) p(\mathcal{D}_{t} | \boldsymbol{\theta}) p(\mathcal{D}_{\mathcal{M}, t} | \boldsymbol{\theta})} d\boldsymbol{\theta}\\
&=-\int q_{t}(\boldsymbol{\theta}) \log \frac{q_{t - 1}(\boldsymbol{\theta}) p(\mathcal{D}_{t} | \boldsymbol{\theta}) p(\mathcal{D}_{\mathcal{M}, t} | \boldsymbol{\theta})}{q_{t}(\boldsymbol{\theta})} d\boldsymbol{\theta}\\
&=\argmax \int q_{t}(\boldsymbol{\theta}) \log \frac{q_{t - 1}(\boldsymbol{\theta}) p(\mathcal{D}_{t} | \boldsymbol{\theta}) p(\mathcal{D}_{\mathcal{M}, t} | \boldsymbol{\theta})}{q_{t}(\boldsymbol{\theta})} d\boldsymbol{\theta}\\
&= \int q_{t}(\boldsymbol{\theta}) \log  p(\mathcal{D}_{t} | \boldsymbol{\theta}) p(\mathcal{D}_{\mathcal{M}, t} | \boldsymbol{\theta}) d\boldsymbol{\theta}\\& \qquad+ \int {q_{t}(\boldsymbol{\theta})} \log \frac{q_{t-1}(\boldsymbol{\theta})}{q_{t}(\boldsymbol{\theta})}d\boldsymbol{\theta}\\
&= \int q_{t}(\boldsymbol{\theta}) \log  p(\mathcal{D}_{t} | \boldsymbol{\theta})d\boldsymbol{\theta} + \int q_{t}(\boldsymbol{\theta}) \log p(\mathcal{D}_{\mathcal{M}, t} | \boldsymbol{\theta}) d\boldsymbol{\theta}\\  & \qquad\qquad- \int {q_{t}(\boldsymbol{\theta})} \log \frac{q_{t}(\boldsymbol{\theta})}{q_{t-1}(\boldsymbol{\theta})}d\boldsymbol{\theta}\\
&= \mathbb{E}_{\boldsymbol{\theta}\sim q_{t}(\boldsymbol{\theta})}\left[\log p(\mathcal{D}_{t} | \boldsymbol{\theta})\right]+\mathbb{E}_{\boldsymbol{\theta}\sim q_{t}(\boldsymbol{\theta})}\left[\log p(\mathcal{D}_{\mathcal{M}, t} | \boldsymbol{\theta})\right]\\
&\qquad\qquad-\text{KL}\left[q_{t}(\boldsymbol{\theta})||q_{t-1}(\boldsymbol{\theta})\right]\\
&= \mathbb{E}_{\boldsymbol{\theta}\sim q_{t}(\boldsymbol{\theta})}\left[\log p(\mathcal{D}_{t} | \boldsymbol{\theta})\right]+\mathbb{E}_{\boldsymbol{\theta}\sim q_{t}(\boldsymbol{\theta})}\left[\log p(\mathcal{D}_{\mathcal{M}, t} | \boldsymbol{\theta})\right]\\
&\qquad\qquad-\lambda_1 \text{KL}\left[q_{t}(\boldsymbol{\theta})||q_{t-1}(\boldsymbol{\theta})\right]\\
&=\argmax \mathbb{E}_{\boldsymbol{\theta} \sim q_{t}(\boldsymbol{\theta})} \left[ \log{p(y_{t} | \boldsymbol{\theta}, G(\boldsymbol{x}_{t}))} \right]\\ &\qquad\qquad+ \sum_{n = 1}^{N_{1}^{\prime}} \mathbb{E}_{\boldsymbol{\theta} \sim q_{t}(\boldsymbol{\theta})} \left[ \log{p(y_{\mathcal{M}, t}^{(n)} | \boldsymbol{\theta}, \boldsymbol{z}_{\mathcal{M}, t}^{(n)})} \right]\\
& \qquad\qquad\qquad- \lambda_{1} \cdot \text{KL}\left(q_{t}(\boldsymbol{\theta}) \left|\right| q_{t - 1}(\boldsymbol{\theta}) \right)
\end{align*}

\noindent where $\mathcal{D}_{t} = \{ \boldsymbol{d}_{t} \} = \{ ( \boldsymbol{x}_{t}, y_{t} ) \}$, $\mathcal{D}_{\mathcal{M}, t} = \{ \boldsymbol{d}^{(n)}_{\mathcal{M}, t} \}^{N_{1}^{\prime}}_{n = 1} = \left\{ ( \boldsymbol{z}_{\mathcal{M}, t}^{(n)}, y_{\mathcal{M}, t}^{(n)} ) \right \}^{N_{1}^{\prime}}_{n = 1}$, $| \mathcal{D}_{\mathcal{M}, t} | = N_{1}^{\prime} \ll | \mathcal{M} |$, $\mathcal{D}_{\mathcal{M}, t} \subset \mathcal{M}$, and $\lambda_{1}$ is a hyper-parameter.

% For \emph{streaming class-instance} ordering, the base initialization is performed by training the network offline with the samples from randomly chosen $2$ classes. In each incremental step, the network is trained in streaming manner with the samples from randomly selected 2 unobserved classes; however, in this case, (i) samples within a class are temporarily ordered based on different object instances, and (ii) all samples from one class are fed into the network before any sample from the other class. 

% In case of \emph{streaming instance} ordering, $10\%$ randomly selected samples are used for base initialization, and the remaining samples are used to train the network incrementally one sample at a time; in streaming setting, the samples are temporally ordered based on different object instances; specifically, we organize the data-stream by putting temporally ordered $50$ frames of an object instance then we put temporally ordered $50$ frames of the second object instance and so on, in this way after putting $50$ temporally ordered frames from each object instance we put next $50$ temporally ordered frames of the first object instance and so on until all the frames of each object instance has exhausted. 

\end{document}